\documentclass[11pt]{article}

\usepackage[final]{acl}

\usepackage{times}
\usepackage{latexsym}

\usepackage[T1]{fontenc}

\usepackage[utf8]{inputenc}

\usepackage{microtype}

\usepackage{inconsolata}

\usepackage{graphicx}

\usepackage{natbib}  
\usepackage{caption} 
\usepackage{algorithm}
\usepackage{algpseudocode}
\usepackage{epsfig}
\usepackage{amsmath}
\usepackage{amssymb}
\usepackage{booktabs}
\usepackage{multirow}
\usepackage[many]{tcolorbox}
\usepackage{cleveref}
\usepackage{subcaption}
\usepackage{pgfplots}
\pgfplotsset{compat=newest}
\usepackage{pgfplotstable}
\usepackage{bm}
\usepackage{newfloat}
\usepackage{listings}
\usepackage[table]{xcolor}

\newcommand{\name}{Perception Magnifier}
\newcommand{\abb}{PM}
\newcommand{\tbf}[1]{\textcolor{blue}{\textbf{#1}}}

\let\subsubsection\paragraph

\title{Through the Magnifying Glass: Adaptive Perception Magnification for Hallucination-Free VLM Decoding}

\author{Shunqi Mao \qquad Chaoyi Zhang \qquad Weidong Cai \\
  School of Computer Science, The University of Sydney, Australia \\
  \texttt{\{shunqi.mao, chaoyi.zhang, tom.cai\}@sydney.edu.au}}

\begin{document}
\maketitle
\begin{abstract}
Existing vision-language models (VLMs) often suffer from visual hallucination, where the generated responses contain inaccuracies that are not grounded in the visual input. 
Efforts to address this issue without model finetuning primarily mitigate hallucination by contrastively
reducing language biases or amplifying the weights of visual embedding during decoding.
However, these approaches remain limited in their ability to capture fine-grained visual details.
In this work, we propose the \name{} (\abb{}), a novel visual decoding method that 
iteratively isolates relevant visual tokens based on attention and magnifies the corresponding regions,
spurring the model to concentrate on fine-grained visual details during decoding.
By magnifying critical regions while preserving the structural and contextual information at each decoding step,
\abb{} allows the VLM to enhance its scrutiny of the visual input, hence producing more accurate and faithful responses.
Extensive experimental results demonstrate that \abb{} not only achieves superior hallucination mitigation but also enhances language generation while preserving strong reasoning capabilities.
Code can be found at \url{https://github.com/ShunqiM/PM}.
\end{abstract}

\section{Introduction}

The evolution of large vision-language models (VLMs)~\cite{llava, blip3, cogvlm, qwen, gpt4v} has revolutionized the integration of computer vision and natural language processing, enabling seamless interaction between visual and textual modalities. These models effectively translate complex visual inputs into coherent and contextually relevant textual descriptions, driving advancements across a wide range of applications~\cite{visual_gpt, palme, ofa}. Fueled by innovations in model architectures \cite{Desai_2021_CVPR, magma}, training methodologies \cite{clip, llava, instructblip}, and diverse multimodal datasets \cite{laion, cc3m}, large VLMs continue to expand their capabilities and adaptability, solidifying their role as foundational tools in modern AI systems.

Despite their remarkable abilities, VLMs are prone to hallucination, where generated outputs fail to align with the visual input, with inaccuracies or unfaithful content.
While some hallucinations arise from VLMs sharing architectural foundations with large language models (LLMs), thereby inheriting their intrinsic limitations \cite{vlm_hallu_survey}, the integration of visual and language features can also lead to hallucinations due to over-reliance on autoregressive language priors and insufficient visual representation capacity.

\begin{figure}[t]
  \centering
    \includegraphics[width=0.9\linewidth]{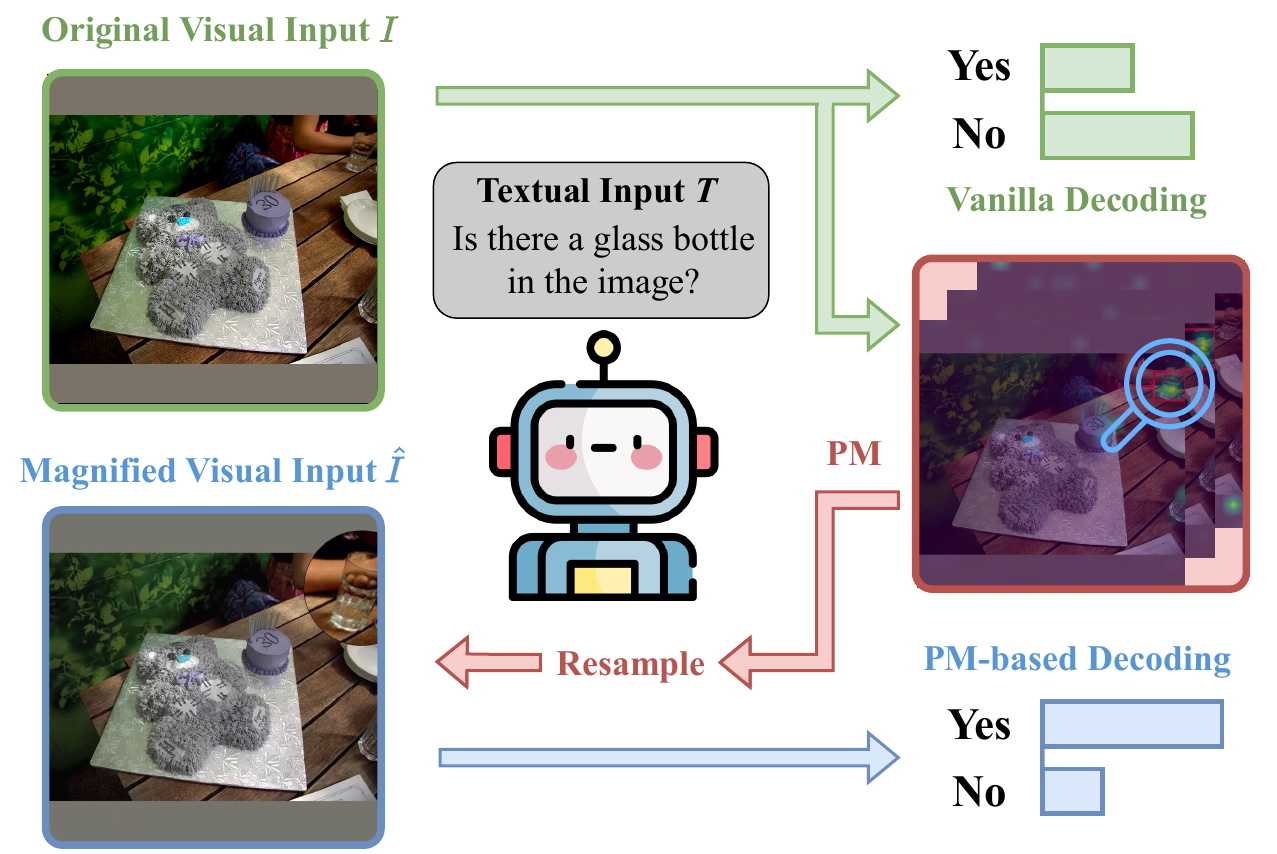}

  \caption{
  We propose \name{} (\abb{}), a hallucination-mitigating decoding framework that
  locates and magnifies key visual regions based on inference attention patterns, enabling feature processing with finer details. 
    For instance, \abb{} locates the attention pattern on the top-right corner, and magnifies the glass bottle to it more discernible to the VLM.
  }
  \label{fig::teaser}\vspace{-3mm}
\end{figure}

Various approaches have been proposed to mitigate hallucination, ranging from training-time techniques, such as debiased datasets \cite{liu2024mitigating, reflective_data_bias} or enlarged visual resolution~\cite{zhai2023halleswitch, monkey}, to inference-time methods such as post-processing for hallucination removal \cite{Woodpecker, lure} and specialized decoding techniques \cite{cd, opera, vcd}. 
Among these, decoding-based approaches have gained particular attention for their ability to improve visual faithfulness without requiring model finetuning.
These methods primarily aim to enhance the contribution of visual information during model inference to mitigate hallucination. This enhancement can be achieved explicitly: by increasing the weight of visual embeddings \cite{pai, ibd}, prompting key regions \cite{api, fine_vp}, or implicitly: by contrastively penalizing outputs that exhibit vague~\cite{vcd, ensemble_vcd}, absent~\cite{vdd, m3id}, or confused visual interpretations~\cite{icd, hio}. 
However, these methods face still limitations in scenarios where insufficient visual resolution causes object misinterpretation.

Inspired by the observation that VLMs allocate marginal attention to many visual tokens during inference and that patch-based architectures struggle with scale-sensitive visual details, we propose the \name{} (\abb{}), a visual decoding approach that adaptively modulates spatial resolution. As shown in \Cref{fig::teaser}, \abb{} magnifies critical regions for higher-resolution processing while spatially compressing rather than discarding less relevant areas, thus enhancing fine-grained detail recognition without compromising structural visual context or intrinsic reasoning ability.

Specifically, we introduce a gradient-free approach to iteratively identify visual tokens with high attention weights across layers, which are most relevant to generation, and aggregate their attention scores into a token-level heatmap. This heatmap is post-processed to enhance coverage and smoothness, then upsampled to construct a pixel-level perception map. Guided by this map, \abb{} refines the visual input by magnifying relevant regions while compressing less critical areas, designed as a resampling process \cite{mag1, mag2}. This dynamic adjustment allows the model to iteratively refine its focus throughout decoding, ensuring improved visual faithfulness without disrupting language reasoning.

\abb{} is broadly applicable to interleaved VLM with unified visual sampling. 
Experiments on hallucination benchmarks along with GPT-4o-assisted assessments in both reference-based and vision-based settings,
demonstrates that \abb{} not only shows superior hallucination mitigation performance at question answering and open-end generation but also preserves the reasoning capabilities of VLMs. Additionally, we provide extensive ablations to analyze the contributions of each component in our method. Our contributions can be summarized as follows:

\begin{itemize} 
    \item We propose the \name{} (\abb{}), a decoding technique that mitigates hallucination by adaptively magnifying visual regions of interest during decoding.
    \item We introduce a novel attention grounding method that iteratively aggregates attentions across layers to generate a perception map, which guides region magnification during decoding.
    \item We conduct experiments with extensive evaluations, demonstrating the effectiveness of \abb{} in mitigating hallucination while preserving reasoning capabilities.
\end{itemize}

\section{Related Work}

\subsection{Hallucination in Vision-Language Models}
Modern vision-language models (VLMs) typically connect vision encoders with large language models (LLMs) via either cross-attention mechanisms~\cite{blip2, instructblip, flamingo} or projector-based token concatenation~\cite{llava, qwen, fromage}, with the latter dominating recent designs~\cite{llava1.5, cogvlm2, blip3} due to its simplicity and efficiency. While such designs align visual and textual modalities effectively, hallucination remains a persistent issue, arising not only from inherited LLM limitations~\cite{vlm_hallu_survey} but also from modality imbalance and vision encoder constraints.

To mitigate hallucination, prior efforts enhance visual grounding by prompting input images~\cite{api, fine_vp, gizdov2024seeing}, boosting visual token influence~\cite{vdd, ibd, pai}, or improving vision module capacity~\cite{zhai2023halleswitch, llava1.5, internvl}. Data-centric strategies include debiasing~\cite{esreal, liu2024mitigating, reflective_data_bias, clip_dpo, lessismore}, contrastive training~\cite{Jiang_2024_CVPR}, preference optimization \cite{xie2024v}, evaluation techniques~\cite{Kaul_2024_CVPR, Guan_2024_CVPR, HaloQuest, pope, nope, faithscore, he2025evaluating}, and post-hoc filtering~\cite{Woodpecker, lure}. 
While these approaches tackle hallucination from multiple angles, few methods focus on enhancing visual recognition capabilities beyond increasing resolution.
We propose a decoding strategy that magnifies semantically important regions without disrupting spatial coherence, improving perception during generation.

\subsection{Hallucination Mitigation at Inference}

Some decoding-end approaches reduce hallucination by modifying next-token generation. Contrastive decoding (CD) \cite{cd} amplifies expert model logits while suppressing outputs from a biased source, typically a smaller or noisier model. 
Subsequent works further extend CD by instantiating the biased source with alternative logits, such as early-layer logits \cite{dola} or context-unaware logits \cite{shi2024trusting}.
OPERA~\cite{opera} penalizes beam outputs that heavily attend to generic aggregation tokens.
VCD~\cite{vcd} adapts CD for VLMs by defining biased outputs from noisy image inputs. Later works extend this by deriving biased outputs by removing image inputs~\cite{vdd, m3id} or enhance focus on textual feature \cite{only}, masking key visual tokens~\cite{sid}, or synthesizing biased outputs through finetuning~\cite{hio} or prompting~\cite{icd}. Others enhance expert logits by upweighting visual tokens~\cite{pai, ibd, chen2025mixture} or manipulating attentions patterns~\cite{su2025activation, prabhakaran2025vade}. Some works explore effectively leveraging multiple outputs path \cite{look_compare, ensemble_vcd}.
While these decoding strategies mitigate hallucination by penalizing visually inconsistent outputs, over-suppressing biased signals may hinder coherence and reasoning. Alternatively, some inference-time methods enhance visual grounding by applying masks or bounding boxes~\cite{api, fine_vp, som, halc}, but still struggle to recognize fine-grained details. ViCrop~\cite{vcrop} improves upon this by cropping and resizing key regions as auxiliary inputs. However, it disrupts spatial structure and may impair tasks requiring holistic understanding, and doubles visual tokens' compute. In contrast, \abb{} magnifies semantically important regions while compressing less relevant areas, preserving spatial integrity and supporting both visual grounding and coherent generation.

\begin{figure*}[!htb]
  \centering
    \includegraphics[width=0.9\linewidth]{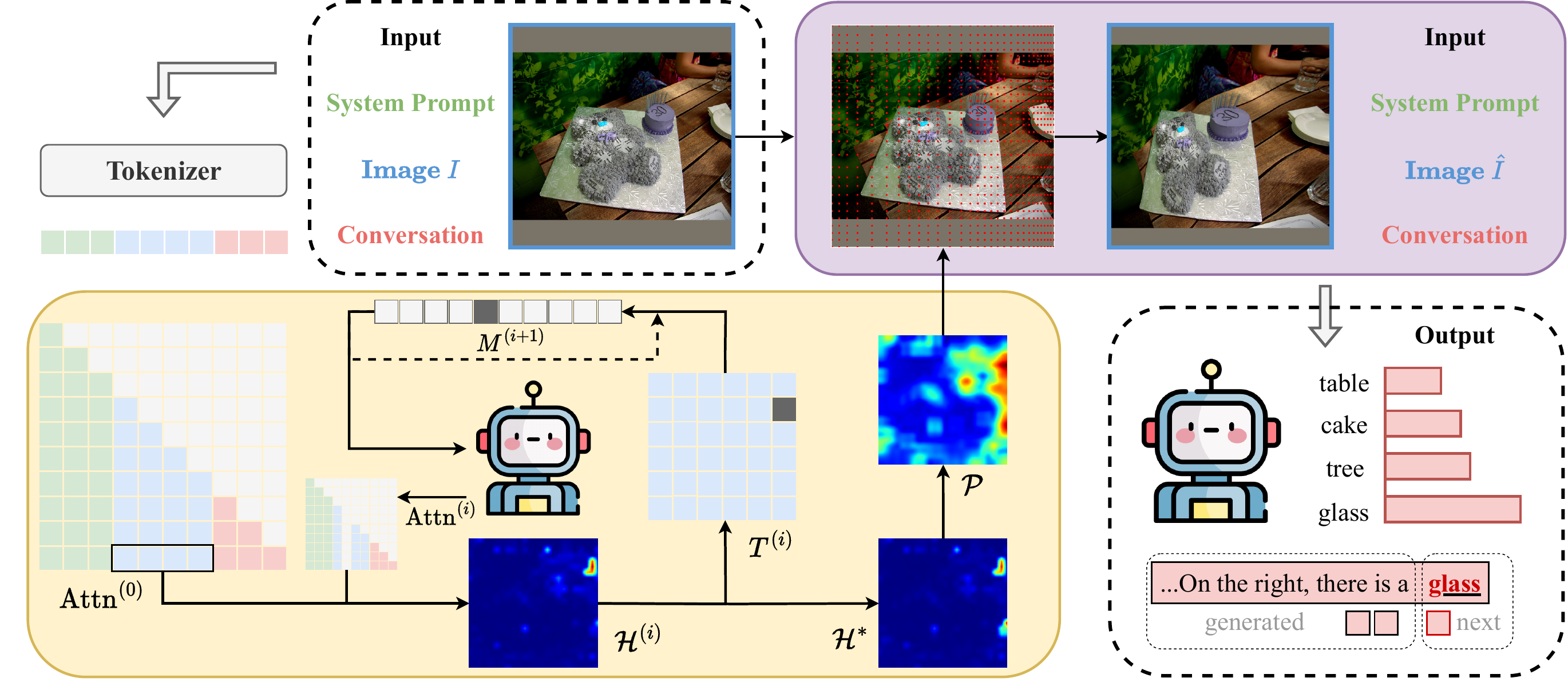}

    \caption{
  Structural diagram of the proposed \abb{} method. At each decoding step, we aggregate multi-layer attentions into a token-level heatmap $\mathcal{H}^{(i)}$. By iteratively masking the most attended tokens and re-forwarding, we collect heatmaps $\{\mathcal{H}^{(i)}\}$ and combine them into $\mathcal{H^*}$ once the process ends, and post-processing $\mathcal{H^*}$ to obtain $\mathcal{P}$ (yellow block). The magnifier resamples the visual input according to  $\mathcal{P}$, producing $\hat{I}$ (purple block). The VLM can then generate the next token with enhanced focus on critical regions with finer details.
  }
  \label{fig::main}
\end{figure*}

\section{Methods}

In this paper, we propose the \name{} (\abb{}), a novel approach that iteratively identifies and enhances key regions of the input image to refine visual input for less hallucinative token generation in VLMs. As illustrated in \Cref{fig::main}, \abb{} iteratively identifies highly attended visual tokens, blocks them, and re-forwards the sequence until no tokens receive significant attention. The identified tokens are then aggregated based on their attention scores to construct a perception map $\mathcal{P}$. This map is used to magnify the original image $I$ via a sampling process, $\hat{I}^i = \mathcal{T}(I, \mathcal{P})$. The refined image $\hat{I}^i$ then serves as the refined visual input for generating the next-step output logits, $O^{i+1} = \text{VLM}(\hat{I}^i, y_{\leq i})$, where $y_{\leq i}$ denotes the previously generated tokens.
The following sections detail the perception map construction, the iterative refinement process for improving the coverage of perception map, and the attention-based magnification processed designed in the proposed \abb{} method.

\subsection{Perception Map Construction} \label{sec::pm_construction}

In this section, we first introduce a simple method to construct a basic perception map to guide the magnifier. 
This method applies to mainstream interleaved VLMs, where patchified visual embeddings are integrated with textual tokens into a unified input sequence processed by the VLM.

The transformer processes  the input sequence via self-attention. 
For a single attention head, the attention weight from token $x_i$ to $x_j$ is computed as $ \mathrm{Attn}_{i,:} = \text{Softmax} \left( Q_i K_j^\top / \sqrt{d_k} \right) $, where $Q_i$ and $K_j$ are the query and key vectors of tokens $x_i$ and $x_j$, respectively, and $d_k$ is the key dimension. The corresponding value $V_j$ is then weighted by $\mathrm{Attn}_{i,j}$ to obtain the hidden representation of $x_i$: $z_i = \sum_{j} \mathrm{Attn}_{i,j} \, V_j$. 
In practice, transformers stack multiple layers and heads, producing attention weights across different levels. These attention patterns can be represented as a multi-dimensional tensor: $\mathrm{Attn} \in \mathbb{R}^{N_{l}, N_{h}, H, W}$, where $N_{l}$ and $N_{h}$ denote the number of layers and attention heads, respectively, and $H \times W$ represents the spatial dimensions of visual tokens, which can be reshaped into a 2D feature map.
\abb{} utilizes the $\mathrm{Attn}$ cached during forward to construct a token-level heatmap, highlighting the most relevant visual regions based on the model attention.

Empirically, we find that attention maps from middle layers more accurately localize relevant objects than those from final layers. This aligns with recent findings showing that deep networks tend to aggregate and process visual information in early layers~\cite{registers}, and that intermediate features can already support strong performance~\cite{dola}.
Hence, we construct the perception map by aggregating self-attention matrices from middle-to-late layers $l \ge \mathcal{L}$ on the last generated token, and applying a max pooling across all heads $h$ to highlight the impact of more visually important regions. Formally, the token-level heatmap $\mathcal{H}$ can be obtained by: 
\begin{equation} 
    \mathcal{H} = \sum_{l=\mathcal{L}}^{N_l} \max_{h \in {1, \ldots, N_h}} \mathrm{Attn}_{l,h}. 
\end{equation} 
Such $\mathcal{H} \in \mathbb{R}^{H, W}$ reflect the intermediate reasoning of VLM and highlight the salient regions identified.

\subsubsection{Post-Processing to Pixel Space.}\label{sec:post}
The basic perception map constructed above without iterative refinement could already be upsampled to guide the magnifier. 
As post-processing before up-sampling the token-level heatmap $\mathcal{H}$ to pixel-level perception map, we first refine the heatmap with methods similar to~\cite{api} to obtain a more smooth and well-distributed attention distribution for better magnification quality.
Firstly, we normalize $\mathcal{H}$ by scaling it to unit interval, obtaining $\mathcal{H}^{\prime}$. Then, we amplify the variance of the normalized map by a scaling coefficient of $\alpha$, and then compress its values back into the range $[0, 1]$ using a sigmoid function, resulting in $\mathcal{H}_{\text{proc}}$:

\begin{equation}
\mathcal{H}_{\text{proc}} = \text{sigmoid}\!\left(\alpha \cdot\,\frac{\mathcal{H}^{\prime} - \mu(\mathcal{H}^{\prime})}{\sigma(\mathcal{H}^{\prime})}\right).
\end{equation}

Finally, we apply a uniform smoothing filter with a kernel size of $k$ to the enhanced heatmap, followed by bilinear upsampling to obtain the pixel-level perception map $\mathcal{P}$. 
As shown in \Cref{fig::main}, this post-processing step refines the heatmap by amplifying low-magnitude attention values, ensuring that small but potentially important regions are also considered for magnification.

\subsection{Iterative Refinement}

While the basic perception map already provides a simple but effective guide for magnification, it often misses some key objects' regions when they appear multiple times. This stems from how deep vision models compress fine-grained features into a small set of tokens, commonly referred to as information registers~\cite{registers, mecha}. VLMs concentrate attention to these tokens and reduce sensitivity to semantically similar but spatially distinct regions.

To improve coverage of the perception map, we introduce an iterative refinement mechanism that progressively reveals overlooked areas by suppressing highly attended tokens from prior passes. 
Nonetheless, directly masking key visual regions at the image level would require fine-grained localization and re-tokenization of visual tokens, making it computationally expensive. Instead, we perform attention blocking at the token level (Alg.~\ref{alg:refinement}, lines 4–5),
which still blocks aggregated tokens, efficiently mitigates above challenges, and integrates easily into existing VLM pipelines.

At each iteration, we forward the input with an attention mask $M^{(i)}$, initialized as a vector of ones (line 1), and extract the token-level heatmap $\mathcal{H}^{(i)}$ (line 5). To identify visually dominant tokens $T^{(i)}$, we apply a simple 2-means clustering algorithm that partitions attention values into two groups (line 7), and select tokens in the one with the higher centroid as $T^{(0)}$. This leverages the observation that relevant tokens typically receive significantly higher attention than others. We then update the mask to exclude these tokens in the next iteration (line 11), encouraging the model to reallocate attention to previously under-attended regions.

The process continues until the total attention in the new heatmap drops below a threshold $\beta$ or a maximum number of iterations is reached (lines 8–10), indicating that no additional high-attention regions remain. We aggregate attention maps across all iterations and normalize them to obtain the final refined heatmap $\mathcal{H}^*$ (lines 13), which is then post-processed to produce the pixel-level perception map $\mathcal{P}$ (line 14). This refined map captures a broader and more accurate set of visually relevant regions, offering better guidance for magnification-enhanced decoding in different vision-language tasks.

\begin{algorithm}[t]
\caption{Iterative Refinement}
\label{alg:refinement}
\begin{algorithmic}[1]
\Require Tokenized input $(I, Q)$, Visual Language Model (VLM), threshold $\beta$, max iterations $T_{\max}$
\State Initialize attention mask $M^{(0)} \gets \mathbf{1}$
\State Initialize heatmap $\mathcal{H}^* \gets \mathbf{0}$, count $\mathcal{C} \gets \mathbf{0}$
\For{$i = 0$ to $T_{\max}$}
    \State $O^{(i)}, \text{Attn}^{(i)} \gets \text{VLM}(I, Q, M^{(i)})$
    \State $\mathcal{H}^{(i)} = \sum_{l \in \mathcal{L}} \max_{h} \mathrm{Attn}^{(i)}_{l,h}$
    
    \State Update $\mathcal{H}^* \gets \mathcal{H}^* + \mathcal{H}^{(i)}$, $\mathcal{C} \gets \mathcal{C} + \mathbf{1}_{M^{(i)}}$
    \State Identify $T^{(i)} \gets \text{Cluster}(\mathcal{H}^{(i)})[0] $
    \If{$\sum \mathcal{H}^{(i)} < \beta$}
        \State \textbf{break}
    \EndIf
    \State Mask out $T^{(i)}$: $M^{(i+1)} \gets M^{(i)} \cap \neg T^{(i)}$
\EndFor
\State Normalize heatmap: $\mathcal{H}^* \gets \mathcal{H}^* \div \mathcal{C}$
\State Post-process $\mathcal{H}^*$ to obtain pixel-level map $\mathcal{P}$
\State \Return Refined perception map $\mathcal{P}$
\end{algorithmic}
\end{algorithm}

\subsection{Attention-Based Magnification}

Inspired by \cite{mag1}, we model the perception magnifier as an attention-based sampling process. The perception map $\mathcal{P}$ is treated as a mass function, where regions with higher perception values are more likely to be sampled. 
Specifically, we first decompose $\mathcal{P}$ into marginal-like distributions along the horizontal and vertical dimensions and approximate their cumulative distributions $\mathcal{F}_x(n)$ and $\mathcal{F}_y(n)$ as:
\begin{equation}
    \begin{split}
        \mathcal{F}_x(n) := \sum_{j=1}^n \max_{1 \le i \le w}\mathcal{P}_{i,j},
        \\
        \mathcal{F}_y(n) := \sum_{i=1}^n \max_{1 \le j \le h}\mathcal{P}_{i,j},
    \end{split}
\end{equation}
where $w$ and $h$ denote the width and height of the perception map, respectively.
Next, we apply inverse transform sampling to remap the original image coordinates to resampled coordinates, ensuring structure-preserving magnification along both dimensions. The resulting image is constructed as:
\begin{equation}
    \hat{I} = \mathcal{T}(I, \mathcal{P})_{i,j} = \text{Interp} \big(I, \mathcal{F}_x^{-1}(i), \mathcal{F}_y^{-1}(j) \big),
\end{equation}
where $\mathcal{F}^{-1}(\cdot)$ denotes the inverse function of $\mathcal{F}(\cdot)$, which remaps uniformly spaced output coordinates to a perception-aware warped space. To ensure smooth spatial mapping, we use bilinear interpolation as the $\text{Interp}$ method in the equation, balancing efficiency and effectiveness, though other interpolation methods could also be applied.  
The magnified image $\hat{I}$ replaces the original visual input $I$ and is forwarded through the VLM with the same textual input to obtain updated model output logits $\hat{O}$. While perception-magnified logits $\hat{O}$ can be combined proportionally or contrastively with the original model output $O$ for next-token generation, we directly sample from $\hat{O}$ after the softmax function, as it already provides an enhanced representation of the visual content.  
The entire perception magnification process is applied at each auto-regressive decoding step, enabling the model to adaptively focus on different regions of the visual input while preserving structural integrity, hence enhancing perception ability and mitigating hallucination.

\begin{table*}[t!]
\centering

\small
\setlength{\tabcolsep}{5pt}
\begin{tabular}{lccccc|c|cccc}
\toprule
\multicolumn{1}{c}{\multirow{2}{*}{}} & \multicolumn{2}{c}{\textbf{Object-level}} & \multicolumn{2}{c}{\textbf{Attribute-level}} & \textbf{MME} & \textbf{MME} & \multicolumn{4}{c}{\textbf{POPE Accuracy}} \\
 & \textbf{Existence} & \textbf{Count} & \textbf{Position} & \textbf{Color} & \textbf{Percep*} & \textbf{Cognition} & \textbf{AOKVQA} & \textbf{GQA} & \textbf{COCO} & \textbf{AVG*} \\
\midrule
Greedy       & 195.00 & 143.33 & 128.33 & 163.33 & 630.00 & \textbf{357.86} & 84.10 & 84.39 & 85.29 & 84.59 \\
OPERA        & 195.00 & 143.33 & 128.33 & 163.33 & 630.00 & 310.36 & 84.09 & 84.43 & 85.29 & 84.60 \\
VCD          & 190.00 & 143.33 & 120.00 & 155.00 & 608.33 & 327.50 & 84.47 & 83.13 & 85.61 & 84.40 \\
ICD          & 190.00 & 148.33 & 123.33 & 153.33 & 615.00 & 348.21 & 84.23 & 84.89 & 84.92 & 84.68 \\
SID          & 195.00 & 155.00 & 126.67 & 160.00 & 636.67 & 327.86 & 84.39 & 84.17 & 85.81 & 84.79 \\
VDD          & 190.00 & 140.00 & 133.33 & 165.00 & 628.33 & 322.64 & 86.73 & 85.52 & 86.71 & 86.32 \\
M3ID         & 190.00 & 150.00 & 133.33 & 166.67 & 640.00 & 339.36 & 86.52 & 85.46 & 86.73 & 86.24 \\
PAI          & 190.00 & 155.00 & 133.33 & 170.00 & 648.33 & 353.93 & 85.32 & 83.43 & 86.66 & 85.14 \\
IBD          & 190.00 & 160.00 & 133.33 & 170.00 & \underline{653.33} & 352.14 & 86.37 & 84.80 & 86.65 & 85.94 \\
API-C        & 190.00 & 150.00 & 128.33 & 165.00 & 633.33 & 317.50 & 86.87 & 85.04 & 87.31 & \underline{86.41} \\
API-L        & 190.00 & 150.00 & 133.33 & 168.33 & 641.67 & 325.00 & 86.52 & 84.30 & 86.60 & 85.81 \\
ViCrop-R     & 190.00 & 163.33 & 105.00 & 175.00 & 633.33 & 335.00 & 85.53 & 85.19 & 87.43 & 86.05 \\
ViCrop-G     & 195.00 & 153.33 & 96.67  & 175.00 & 620.00 & 350.29 & 86.09 & 85.18 & 87.31 & 86.19 \\
\rowcolor{gray!15}\abb{} (Ours) & {195.00} & {175.00} & {138.33} & {175.00} & \textbf{683.33} & \underline{355.36} & {86.94} & {85.46} & {87.68} & \textbf{86.70} \\
\bottomrule
\end{tabular}

\caption{Evaluation results for MME Perception Scores (left), Cognition Scores (mid), and POPE Accuracy (right). Columns marked with * indicate the overall hallucination scores for the benchmark. The result for BEAM is omitted for its equivalence with OPERA on single-token QA.  API and ViCrop each propose multiple variants, denoted by postfixes (e.g., -C or -L), with details provided in \Cref{sec::sup_base}. 
}\label{tab:main}

\end{table*}

\begin{figure}
    \centering
    \includegraphics[width=\linewidth]{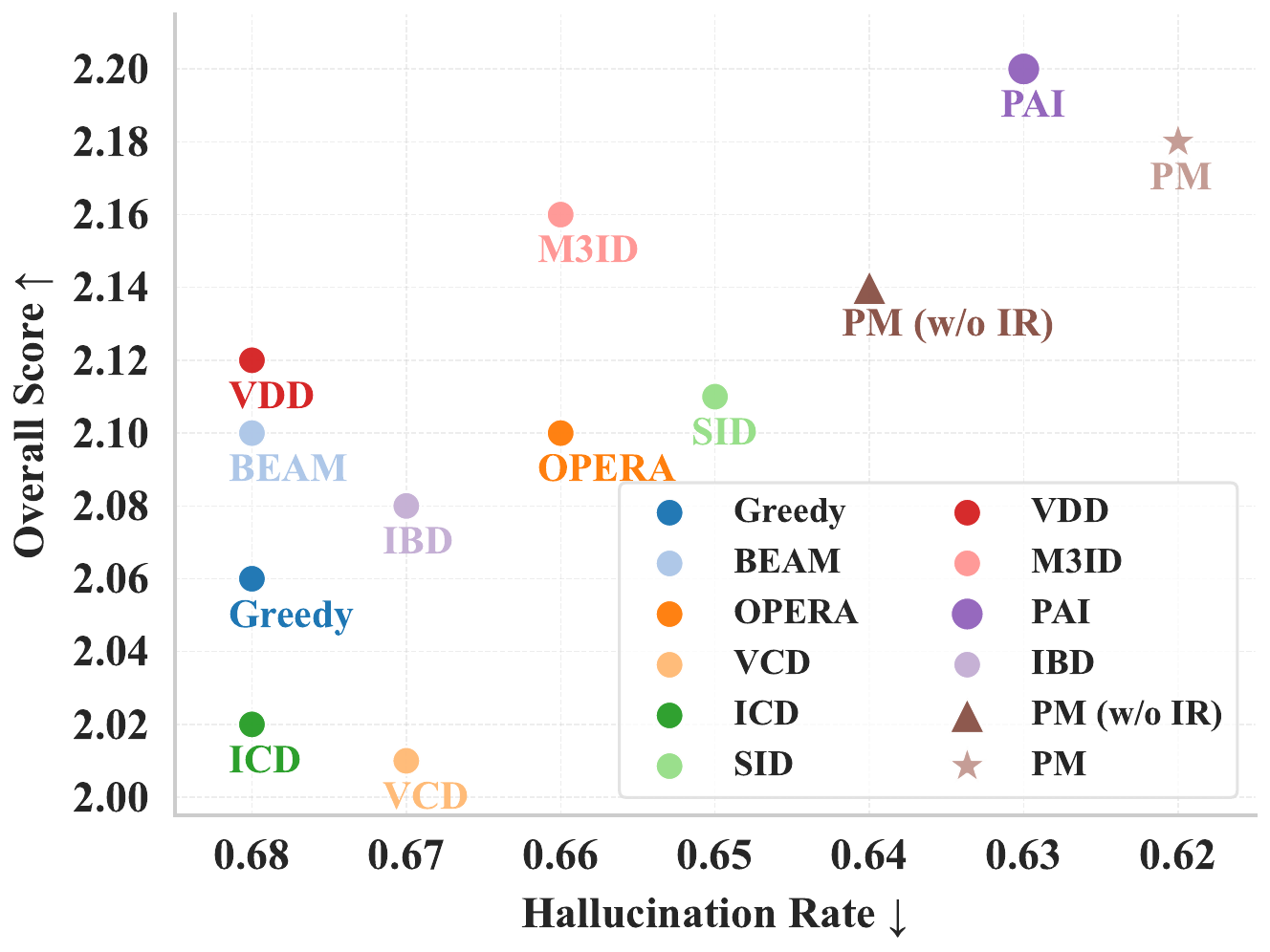}
    \caption{MMHal-Bench comparison of hallucination mitigation decoding methods. Each point represents one method, with lower hallucination rate and higher overall score indicating better performance.}
    \label{fig:mmhal_results}
\end{figure}

\section{Experiments and Results}

\subsubsection{Datasets and Baselines.} We evaluate our method on four benchmarks: MME~\cite{mme}, POPE~\cite{pope}, {MMHal-Bench} \cite{sun2024aligning} and LLaVA-Bench~\cite{llava}, all commonly used for assessing multimodal hallucination. For comparison, we include various decoding or inference strategies, including greedy decoding, VCD~\cite{vcd}, VDD~\cite{vdd}, M3ID \cite{m3id}, IBD~\cite{ibd}, PAI~\cite{pai}, ICD~\cite{icd}, SID~\cite{sid}, beam search~\cite{beam}, OPERA~\cite{opera}, API~\cite{api}, and ViCrop \cite{vcrop}. Additional methods such as \cite{lessismore,shi2024trusting}, which rely on contextual information or EOS decisions, are less relevant to visual hallucination and are therefore excluded. Further details are provided in \Cref{sec::sup_data,sec::sup_base}.

\subsubsection{Implementation Details.} 
We conduct our experiments using LLaVA-1.5 7B~\cite{llava1.5}. We start aggregating attentions from layer $\mathcal{L} = 12$, with scaling coefficient and kernel size $\alpha = 10$, $k=3$, and attention threshold for iterative refinement $\beta = 0.3$.

\subsection{Quantitative Analysis}

Table~\ref{tab:main}
reports results on MME and POPE. POPE centers on existence-based queries, where most methods benefit from reinforcing visual token importance. MME is more challenging, with its perception subset targeting hallucination and its cognition subset requiring higher-level reasoning. While our method focuses on mitigating hallucination, we report cognition results for completeness, noting that most baselines underperform on this subset 
as modifications on the embedding and logit could harm the VLM reasoning.

Among baselines, CD-based methods suppress hallucinated logits but often fail when visual detail remains ambiguous across both logits. Embedding-level methods such as PAI and IBD improve perception by reinforcing visual features across multiple layers. However, they may still struggle when relevant regions are too small or diffuse to be effectively represented in the ViT backbone.
API suppresses irrelevant regions via masking, but does not enhance salient visual content. ViCrop isolates key areas via cropping, which may hinder spatial understanding due to loss of context and potential ambiguity from dual-image inputs, 
resulting in limited MME gains.
In contrast, our method magnifies the relevant regions while preserving surrounding context and internal states of the VLM. This preserves the embedding distribution and enhances fine-grained recognition without compromising spatial coherence or reasoning, leading to state-of-the-art results across both benchmarks.

We further evaluate on MMHal-Bench in Figure~\ref{fig:mmhal_results}, which provides a complementary setting for hallucination assessment in open-end generation. On this benchmark, \abb{} is competitive with the strongest baseline in overall score while achieving the lowest hallucination rate among all compared methods. We attribute the relatively modest gain in overall score to the design of MMHal-Bench, where many questions concern attributes of the primary object in the scene. In such cases, the benefit of magnifying small or cluttered regions is naturally less pronounced. Nevertheless, \abb{} still consistently reduces hallucinations, and the full model further improves over \abb{} w/o IR on both overall score and hallucination rate, supporting the effectiveness of iterative refinement.

In addition, all methods exhibit varying degrees of degradation on the MME cognition subset, whereas ours does not. A detailed analysis of how and why different methods affect the cognitive ability is provided in \Cref{sec::cognition}.

\definecolor{forestgreen}{RGB}{34, 139, 34}
\begin{figure*}[t!]
    \centering

    \begin{minipage}[t]{0.18\textwidth}
    \vspace{-2.6cm}
    
    \begin{tabular}{@{}l@{\hspace{2pt}}r@{}} 
    \textbf{Question:} \\
    Is there a chair \\ in the image? \vspace{0.2cm} \\
    
        Greedy: & \textcolor{red}{\textbf{No}} \\ 
        \abb{}: & \textcolor{forestgreen}{\textbf{Yes}}
    \end{tabular}
\end{minipage}
    \begin{minipage}[t]{0.18\textwidth}
        \centering
        \includegraphics[width=\textwidth]{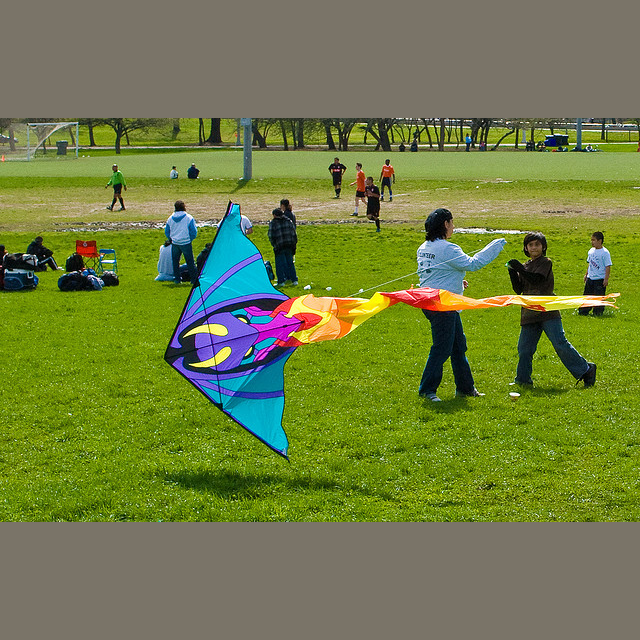} 
    \end{minipage}
    \begin{minipage}[t]{0.18\textwidth}
        \centering
        \includegraphics[width=\textwidth]{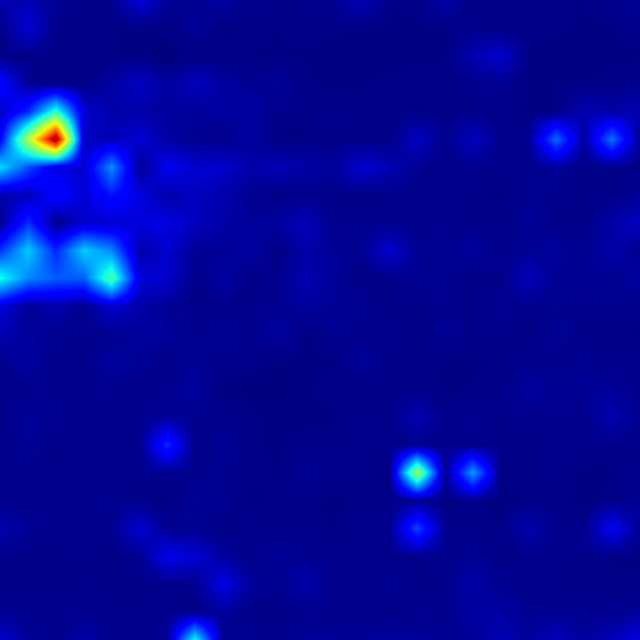} 
    \end{minipage}
    \begin{minipage}[t]{0.18\textwidth}
        \centering
        \includegraphics[width=\textwidth]{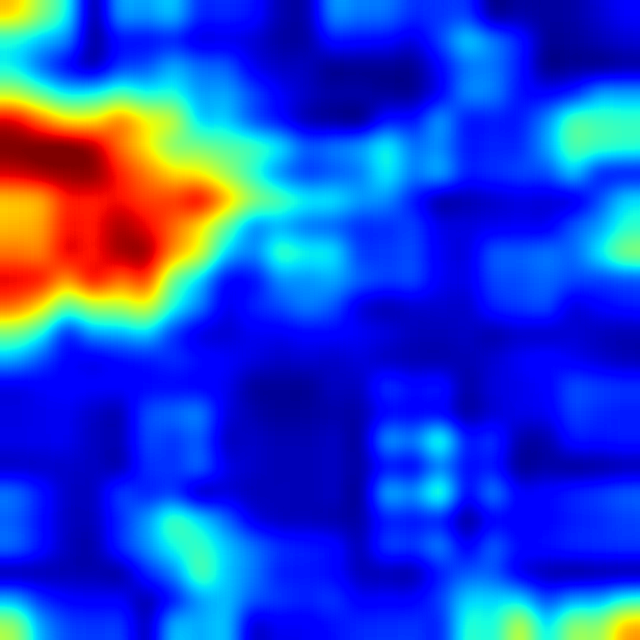}
    \end{minipage}
    \begin{minipage}[t]{0.18\textwidth}
        \centering
        \includegraphics[width=\textwidth]{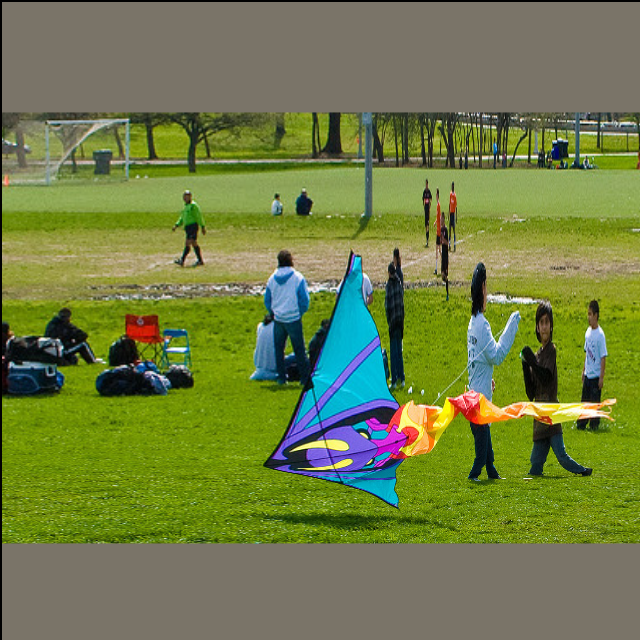}
    \end{minipage}

    \vspace{0.1cm}

    \begin{minipage}[t]{0.18\textwidth}
        \vspace{-2.6cm}
        
    \begin{tabular}{@{}l@{\hspace{2pt}}r@{}} 
    \textbf{Question:} \\
    Is there a tree \\ in the image? \vspace{0.2cm}\\
        Greedy: & \textcolor{red}{\textbf{Yes}} \\ 
        \abb{}: & \textcolor{forestgreen}{\textbf{No}}
    \end{tabular}
    \end{minipage}
    \begin{minipage}[t]{0.18\textwidth}
        \centering
        \includegraphics[width=\textwidth]{samples/padded_ori_819.png} %
    \end{minipage}
    \begin{minipage}[t]{0.18\textwidth}
        \centering
        \includegraphics[width=\textwidth]{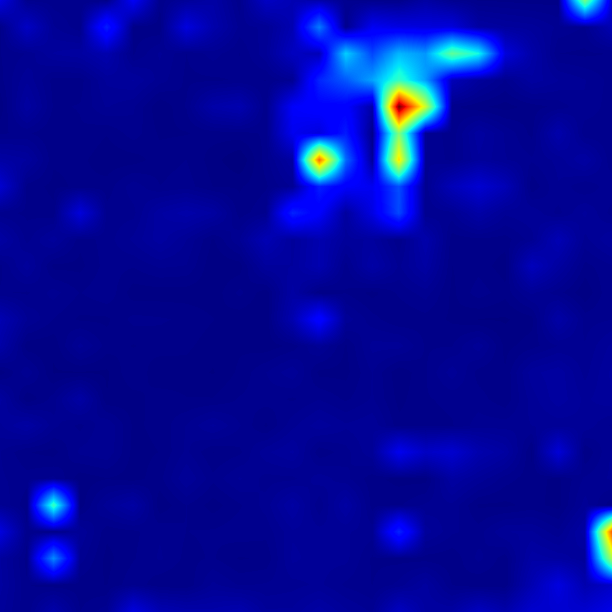} 
    \end{minipage}
    \begin{minipage}[t]{0.18\textwidth}
        \centering
        \includegraphics[width=\textwidth]{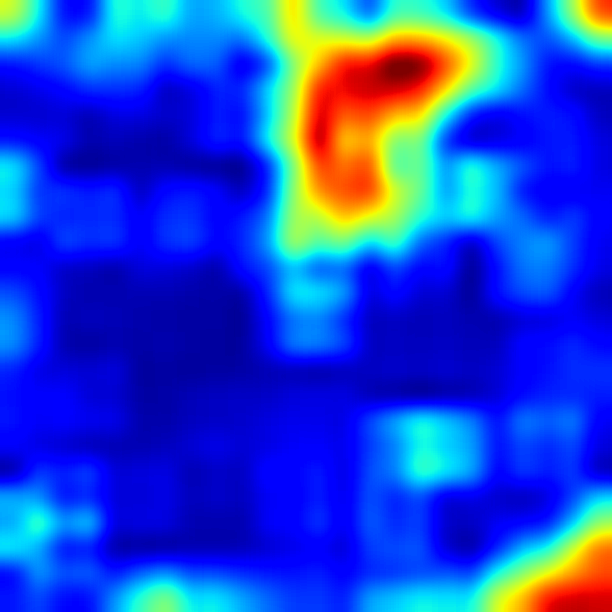}
    \end{minipage}
    \begin{minipage}[t]{0.18\textwidth}
        \centering
        \includegraphics[width=\textwidth]{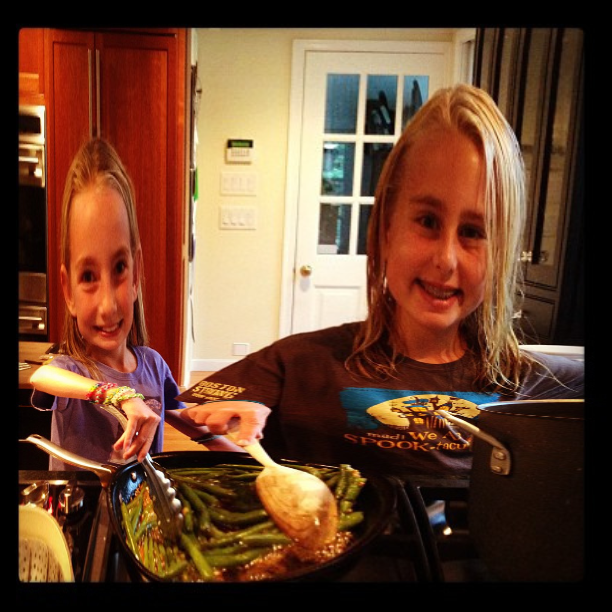}
    \end{minipage}

    \caption{Qualitative examples showing the input question, the original image, the raw and processed perception map $\mathcal{H}^*$ and $\mathcal{P}$, and the perception-magnified image, arranged from left to right. The answers generated by greedy decoding and \abb{} are compared, with correct answers highlighted in green and incorrect ones in red.}\label{fig:qualitative}
\end{figure*}
\subsection{GPT-4o Assisted Evaluation}

\begin{table}[t]
\centering
\scriptsize
\begin{tabular}{lccc|ccc}
\toprule
\multicolumn{1}{p{3mm}}{\multirow{1}{*}{LLaVA-Bench}} & \multicolumn{3}{c|}{\textbf{COCO}} & \multicolumn{3}{c}{\textbf{In-the-Wild}}  \\                \multicolumn{1}{c}{}      &     \textbf{Ref} $\uparrow$    & \textbf{Acc}  $\uparrow$   & \textbf{Det} $\uparrow$     & \textbf{Ref} $\uparrow$    & \textbf{Acc} $\uparrow$   & \textbf{Det}  $\uparrow$     \\
\midrule
Greedy & 81.15     & 92.69  & 95.76 & 56.94 & 71.83 & 79.95 \\
BEAM & 83.17&  94.75 & 91.60 & 58.94 & 74.71 & 78.25\\
OPERA & 82.99 & 96.12 & 93.61 & 58.44 & 71.81 & 77.92 \\
VCD & 82.00&  92.21 & 94.39 & 57.17 & 69.4 & 81.22 \\
ICD  & 83.29 & 93.91 & 95.46 & 54.43 & 66.89 & 74.81 \\
SID  & 81.10 & \underline{97.35} & 95.80 & 59.49 & 70.93 &  81.63 \\
VDD  & 81.38 & 95.62 & \textbf{102.75} & {61.60} & \textbf{78.22} & \underline{83.01} \\
M3ID& 83.02 &   96.41 & 95.78 & \underline{61.88} & 77.49 & 82.83  \\
PAI& 79.04 &   92.06 & 90.59 & 52.29 & 69.74 & 72.49 \\
IBD & \underline{85.29} &  95.73 & 95.45 & 61.59 & \underline{77.91} & 82.86 \\
\rowcolor{gray!15}\abb{} (Ours) & \textbf{85.76} & \textbf{97.74} &  \underline{95.80} &\textbf{62.46}  & {77.21}  & \textbf{84.50} \\
\bottomrule
\end{tabular} 
\caption{Results on LLaVA-Bench (\%). Ref denotes reference aggrement, while Acc and Det represent accuracy and detailedness in vision-based evaluation.}\label{tab:geval}
\end{table}

Previous studies~\cite{opera, ibd} have demonstrated the reliability of GPT-4-assisted evaluation for hallucination assessment.  
We extend their evaluation protocol and experiment on LLaVA-Bench to benchmark the open-end generation quality.
Overall, GPT-4o scores responses based on reference agreement (\textit{Ref}), visual accuracy (\textit{Acc}), and descriptive detail (\textit{Det}), depending on the evaluation setting.
We omit API and ViCrop as they are not applicable to open-ended generation. Details on the GPT-4o metrics and prompts can be found in \Cref{sec::app_gpt}.

As shown in \Cref{tab:geval}, \abb{} achieves the most competitive overall performance on open-end generation tasks. While its detailedness on COCO and accuracy on the In-the-Wild split are slightly lower than a few methods, \abb{} still demonstrates advantage in hallucination mitigation, particularly in reference-based evaluation. On COCO, the detailedness gap appears to be mainly related to response verbosity rather than clearly inferior visual grounding. On the In-the-Wild split, the relatively weaker gains are mostly observed in cases where answers rely more on language priors or world knowledge than on fine-grained visual details, which are less aligned with the strength of PM. 
These results indicate that even in open-end decoding, where the model does not always need to focus on specific regions at each step, the proposed adaptive perception magnifier can still enhances output quality and effectively reduces hallucinations.

\begin{table}[t]
\centering
\setlength{\tabcolsep}{4pt}
\scriptsize
\begin{tabular}{lccccc}
\toprule
\multicolumn{1}{c}{\multirow{2}{*}{MME-Perception}} 
& \multicolumn{2}{c}{\textbf{Object-level}} 
& \multicolumn{2}{c}{\textbf{Attribute-level}} 
&   \multicolumn{1}{c}{\textbf{Percep}} \\
\multicolumn{1}{c}{} & \textbf{Existence} & \textbf{count} & \textbf{position} & \textbf{color} & \textbf{Total*} \\
\midrule
Greedy            
& 195.00 & 143.33 & 128.33 & 163.33 & 630.00 \\
\abb{} w/o IR \& MLA  
& 195.00 & 155.00 & 120.00 & 170.00 & 640.00 \\
\abb{} w/o MLA       
& 195.00 & 155.00 & 125.00 & 170.00 & 645.00 \\
\abb{} w/o IR        
& 190.00 & 175.00 & 135.00 & 165.00 & 665.00 \\
\rowcolor{gray!15}\abb{}               
& {195.00} & {175.00} & {138.33} & {175.00} & \textbf{683.33} \\
\bottomrule
\end{tabular}
\caption{
Ablations on iterative refinement (IR) and multi-layer aggregation (MLA). Both contribute to improvements compared to their respective baselines.
}
\label{tab:abla}
\end{table}

\subsection{Qualitative Analysis}

Figure~\ref{fig:qualitative}
shows qualitative examples depicting the input question, original image, raw and processed perception map $\mathcal{H}^*$ and $\mathcal{P}$, and the perception-magnified image. The results demonstrate that the proposed method effectively enlarges small objects relevant to the question while preserving coarse-grained structural information. 
In the first example, the VLM initially attends to the top-left region, recognizing the potentially relevant region but still failing to identify the chair. {This suggests that accurate attention does not necessarily imply correct recognition, as local evidence may still be insufficient under limited resolution.} With the magnified input, the chair is presented at a higher spatial resolution, enabling the VLM to correctly recognize it.
This illustrates that accurate attention does not imply recognition: local evidence can be overridden under limited resolution.
With the magnified input, the chair is presented at a higher spatial resolution, enabling the VLM to correctly recognize it. Similarly, in another case, the VLM misidentifies an object behind a door window as a tree. However, the perception-magnified input allows the model to re-examine the object with greater detail, preventing hallucinated predictions.

Overall, the qualitative results suggest that the proposed method enhances the perceptibility of fine-grained visual details in regions critical for next-token generation. By enabling the model to examine these areas at a higher resolution, our approach effectively reduces hallucinations and improves overall model performance.
\subsection{Ablations}

\subsubsection{Iterative Refinement.}

We ablate the effect of iterative refinement in constructing the perception map. As shown in \Cref{tab:abla}, the refined maps consistently outperform their non-iterative counterparts, confirming the benefit of this strategy in improving visual localization. Iterative refinement helps address tokens overlooked from the early-stage aggregation of visual features, resulting in more accurate coverage of relevant regions.
Although some residual leakage from masked tokens remains due to information exchange in the ViT~\cite{vit} visual feature extractor, our approach still achieves decent grounding of key areas with minimal computational overhead, making it an effective and efficient solution for guiding the magnifier.

\subsubsection{Multi-Layer Aggregation.}

We ablate the effect of aggregating attention across multiple layers when constructing the perception map. As shown in \Cref{tab:abla}, multi-layer aggregation yields significantly better performance compared to using attention from a single layer.
While middle-to-deep layers generally focus more on visual content, the most informative attention varies across layers and samples. In many cases, only a subset of intermediate layers exhibits structured visual attention, while others focus primarily on sink tokens~\cite{sink}. Aggregating across layers reduces this variability, ensuring that key visual cues are reliably captured for guiding the magnifier.

\begin{table}[t]
\centering
\scriptsize
\setlength{\tabcolsep}{5pt}
\begin{tabular}{lccccc}
\toprule
\multicolumn{1}{c}{\multirow{1}{*}{MME-}} & \multicolumn{2}{c}{\textbf{Object-level}} & \multicolumn{2}{c}{\textbf{Attribute-level}} & \multicolumn{1}{c}{\textbf{Percep}}\\
\multicolumn{1}{c}{\multirow{1}{*}{Perception}} & \textbf{Existence} & \textbf{Count} & \textbf{Position} & \textbf{Color} & \textbf{Total*} \\
\midrule
\multicolumn{6}{l}{\textit{(a) Perception Map Construction Methods}} \\
 API CLIP     & 190.00 & 155.00 & 131.67 & 170.00 & 646.67 \\
 API LLaVA    & 190.00 & 165.00 & 126.67 & 170.00 & 651.67 \\
 ViCrop-R     & 195.00 & 163.33 & 116.67 & 170.00 & 645.00 \\
 ViCrop-G     & 195.00 & 158.33 & 116.67 & 170.00 & 640.00 \\
 Rel Map      & 195.00 & 175.00 & 130.00 & 155.00 & 655.00 \\
\rowcolor{gray!15}
 \abb{} w/o IR & {190.00} & {175.00} & {135.00} & {165.00} & \textbf{665.00} \\
\midrule
\multicolumn{6}{l}{\textit{(b) Perception-Guided Magnification Methods}} \\
 Blurring      & 185.00 & 158.33 & 111.67 & 175.00 & 630.00 \\
 Bounding Box  & 180.00 & 160.00 & 130.00 & 170.00 & 640.00 \\
 Masking          & 195.00 & 155.00 & 128.33 & 170.00 & 648.33 \\
 ViCrop & 195.00 & 153.33 & 123.33 & 175.00 & 646.67 \\
 DualView & 190.00 &	155.00 &	135.00 &	170.00&	650.00 \\
\rowcolor{gray!15}
 Magnification & {195.00} & {175.00} & {138.33} &{175.00} & \textbf{683.33} \\
\bottomrule
\end{tabular}
\caption{Ablation results across two groups: (a) comparing different methods for constructing perception maps; (b) comparing prompting methods that highlight key regions based on the map. IR denotes iterative refinement.}
\label{tab:abla_combined}
\end{table}

\subsubsection{Comparison with Different Maps.}

Attention-based visualization methods~\cite{map1, map2} have been proposed to improve VLM interpretability. We adapt several representative techniques—including CLIP-based Rel(evance) Maps, API~\cite{api}, and ViCrop~\cite{vcrop}, to construct a map to guide the magnifier, and compare them to our approach w/o IR using identical post-processing.
As shown in \Cref{tab:abla_combined}, our method achieves superior coverage of relevant visual regions. Attribution maps based on CLIP often reflect global similarity rather than token-level saliency, while API and ViCrop variants aggregate attention from a single layer, resulting in lower stability and diluted focus. Backpropagation-based methods (e.g., Rel Map, ViCrop-G) offer fine-grained attribution but contribute little to magnification quality and are computationally expensive.
In contrast, the proposed strategy, even w/o the IR process, preserves critical visual cues with better spatial consistency and minimal overhead, making it well-suited for efficient perception-guided decoding.

\subsubsection{Image Prompting Methods.}

Various methods have been proposed to ``prompt" images to emphasize specific regions~\cite{fine_vp, vcrop}, including adding a bounding box, applying blurring effect, masking out the remaining regions, and cropping out the key area (ViCrop), or append the magnified image, as dual inputs (DualView). We benchmark these methods as alternatives to our magnifier, evaluating their effectiveness in highlighting key visual regions when combined with the constructed perception map. As shown in \Cref{tab:abla_combined}, our magnification approach more effectively leverages the perception map to enhance the perception ability of the VLM.

Concretely, magnification not only highlights important regions but also reveals fine-grained details within limited resolution, allowing the model to capture subtle visual patterns. In contrast, many prompting methods merely emphasize region importance without increasing effective resolution, limiting their impact on fine-grained recognition.
Meanwhile, ViCrop also yields certain improvement, as it also enhances local details with resized key regions. However, ViCrop discards contextual information and confuses the VLM by supplying the additional resized image, leading to reduced effectiveness in tasks such as position-related and counting questions. The DualView methods

\section{Conclusion}

We introduce the \name{} (\abb{}), a visual decoding method that refines image inputs to mitigate hallucinations in vision-language models.
\abb{} adaptively enlarges visually critical regions identified by multi-layer attention iteratively, providing enhanced effective resolution on relevant finer-grained details at each decoding step.
Experiments demonstrate that \abb{} significantly reduces hallucinations at VQA and open-end generation, while preserving reasoning capabilities. 
Future work will focus on optimizing the efficiency for \abb{}, and mitigate shape distortions.

\section*{Limitations}
Although \abb{} effectively mitigates hallucinations, it also introduces several limitations. Region magnification may distort local shapes, reducing reliability in tasks requiring geometric fidelity. The method further incurs computational overhead, as magnification interrupts efficient decoding strategies like KV caching. In addition, applying \abb{} to models with complex token–image mappings can require additional attention alignment mechanisms. Further detailed discussions are available at \Cref{sec::limitations}.


\bibliography{custom}

\appendix
\clearpage
\twocolumn[
\begin{center}
\LARGE\bfseries Through the Magnifying Glass: Adaptive Perception Magnification for Hallucination-Free VLM Decoding \\
\vspace{2mm}
\Large Appendix
\vspace{1em}
\end{center}
]

\begin{table*}[t]
\centering
\small
\setlength{\tabcolsep}{5pt}
\begin{tabular}{lccc|ccc|ccc}
\toprule
\multirow{2}{*}{\textbf{Method}} & \multicolumn{3}{c|}{AOKVQA} & \multicolumn{3}{c|}{GQA} & \multicolumn{3}{c}{COCO} \\
& Adversarial & Popular & Random & Adversarial & Popular & Random & Adversarial & Popular & Random \\
\midrule
Greedy & 78.67 & 84.93 & 88.70 & 80.40 & 83.27 & 89.50 & 83.50 & 85.57 & 86.80 \\
Opera & 78.60 & 84.93 & 88.73 & 80.43 & 83.33 & 89.53 & 83.47 & 85.57 & 86.83 \\
VCD & 79.10 & 84.93 & 89.37 & 79.63 & 81.70 & 88.07 & 83.53 & 85.63 & 87.67 \\
ICD & 79.07 & 85.40 & 88.23 & 80.97 & 84.77 & 88.93 & 83.20 & 85.23 & 86.33 \\
SID & 78.90 & 85.40 & 88.87 & 80.60 & 83.33 & 88.57 & 83.90 & 86.07 & 87.47 \\
VDD & 81.03 & 88.27 & 90.90 & 82.00 & 85.27 & 89.30 & 85.00 & 87.00 & 88.13 \\
M3ID & 80.87 & 87.97 & 90.73 &81.67	&84.93	&89.77 & 85.10 & 86.93 & 88.17 \\
PAI & 79.63 & 86.27 & 90.07 & 79.90 & 82.30 & 88.10 & 85.03 & 86.97 & 87.97 \\
IBD & 80.60 & 87.47 & 91.03 & 81.30 & 84.10 & 89.00 & 84.90 & 86.97 & 88.07 \\
API-C & 81.00 & 88.20 & 91.40 & 81.33 & 84.27 & 89.53 & 85.47 & 87.53 & 88.93 \\
API-L & 80.80 & 87.70 & 91.07 & 81.00 & 83.57 & 88.33 & 84.53 & 86.97 & 88.30 \\
ViCrop-R & 80.00 & 86.37 & 90.23 & 81.60 & 83.83 & 90.13 & 85.60 & 87.70 & 89.00 \\
ViCrop-G & 80.97 & 86.77 & 90.53 & 81.67 & 83.93 & 89.93 & 85.13 & 87.70 & 89.10 \\
\rowcolor{gray!15}
\abb{} (Ours) & {81.07} & {87.77} & {92.00} & {81.73} & {84.37} & {90.30} & {85.87} & {88.00} & {89.17} \\
\bottomrule
\end{tabular}
\caption{Detailed evaluation results for POPE accuracy with LLaVA 1.5.}
\label{tab:full_pm}
\end{table*}

\section{Datasets} \label{sec::sup_data}
We carry out our experiments on three benchmarks popular for examining the perception ability and hallucination of VLMs in English. \\
\textbf{MME}~\cite{mme} benchmark is designed to comprehensively assess both perception and cognition abilities of VLMs across 14 distinct subtasks on 2,370 questions. In our study, we focus primarily on the \textit{Perception} subset of MME following \cite{vcd} to evaluate the effectiveness of our method in controlling hallucinations. 
\\
\textbf{POPE} (Polling-based Object Probing Evaluation) \cite{pope} is a 9 $\times$ 3,000 question benchmark for evaluating object hallucination in VLMs. It formulates hallucination evaluation as a binary classification task by posing yes/no questions about object presence, addressing limitations of CHAIR~\cite{chair}, which relies on exact text matching, and providing a more robust evaluation less sensitive to prompt variations. \\
\textbf{LlaVA-Bench} \cite{llava} is a benchmark designed to evaluate VLMs in real-world visual chat scenarios. It comprises two components: LLaVA-Bench (COCO), utilizing images from the COCO dataset, and LLaVA-Bench (In-the-Wild), a more challenging subset featuring 24 diverse images—such as indoor and outdoor scenes, memes, paintings, and sketches—paired with 60 detailed, manually curated questions. The benchmark covers various vision-language tasks, making it suitable for assessing model generalization. In our study, we employ both COCO and In-the-Wild splits to comprehensively evaluate different methods on hallucination control, cognitive abilities, and performance in open-ended question answering.

We also evaluate our method and baselines on \textbf{MMHal-Bench}~\cite{sun2024aligning}, a multimodal hallucination benchmark consisting of 96 open-ended image-question pairs, spanning diverse 8 question types and 12 object categories. MMHal-Bench adopts a GPT-4-based evaluation protocol that measures both average score and hallucination rate for VLM-generated outputs. We do not include additional benchmarks such as MMMU~\cite{yue2024mmmu} and MathVision~\cite{wang2024measuring}, as they are dominated by chart-based and symbolic problems where failures are primarily due to abstract reasoning rather than hallucination or misperception, and are thus largely orthogonal to our focus.

\section{Baselines} \label{sec::sup_base}
This section summarizes the high-level idea of methods used as comparisons in the experiments. We apply these methods on our benchmarks with the default hyperparameters provided in the original paper. 

\noindent \textbf{Beam Search} explores multiple output candidates by iteratively expanding the most probable sequences and selecting the top-$k$ hypotheses at each step, resulting in improved answer diversity but at the cost of increased computational complexity proportional to the beam size. We evaluated all beam-based methods with a beam size of 4.\\
\textbf{OPERA} \cite{opera} is a beam-search-based method, examining the presence of a hallucinate beam and pruning the beam based on its attention on information aggregation patterns.\\
\textbf{VCD} \cite{vcd} introduces noise into the input image and processes it through the VLM to generate a hallucinative output. By leveraging contrastive decoding (CD) to penalize high-probability predictions that align with the hallucinative output, VCD reduces the likelihood of hallucinations in the final prediction, resulting in more reliable responses. \\
\textbf{ICD} \cite{icd} generates a hallucinative output for CD-based decoding by appending \textit{``You are a confused object detector"} to the input prompt.  \\
\textbf{SID} \cite{sid} induces a hallucinative output for CD-based decoding by identifying key visual tokens in the image and pruning them through forward.  \\
\textbf{VDD} \cite{vdd} mitigates language-induced hallucinations in VLMs by generating a hallucinative output through the removal or replacement of visual tokens with \texttt{<unk>} tokens.  \\
\textbf{M3ID} \cite{m3id} is similar to VDD but with weighted strength of incorporating contrastive corrections based on the text-image mutual information.
\\
\textbf{PAI} \cite{pai} enhances visual faithfulness by increasing the weights of visual tokens through attention mechanisms while penalizing outputs generated without visual inputs using CD. 
\\
\textbf{IBD} \cite{ibd} also biases the output toward visual inputs by adding visual attentions and refines it against the original output using CD. Additionally, it adaptively adjusts the contrastive update strength at each decoding step based on contextual needs.  
\\
\textbf{API} \cite{api} highlights key visual regions by overlaying a gray-level mask on the input image. The mask is guided by a heatmap derived from image-text associations using a CLIP model (-C) or a LLaVA model (-L).  
\\
\textbf{ViCrop} \cite{vcrop} estimates a bounding box around the key region in the image and generates a new input by cropping and resizing this region. Three approaches are proposed for bounding box estimation with different level of reliance on gradient information: relative attention (-R), gradient attention (-G), and pure gradient (-P). The pure-gradient approach is not benchmarked due to its higher GPU memory consumption, which necessitates lower model precision and results in degraded performance, making it unsuitable for a fair comparison.

\section{Computing Infrastructure}
All experiments were conducted on a single machine equipped with an NVIDIA GeForce RTX 3090 GPU and an Intel(R) Core(TM) i7-10700 CPU. The system runs Ubuntu 22.04 LTS and uses CUDA version 12.1 for GPU acceleration. Experiments were implemented in Python using PyTorch (v2.4.0) and Transformers (v4.37.2). Experiments on all benchmarks take around 24 GPU hours in total.

\section{License of Artifacts}
We use the following resources in accordance with their respective licenses and terms: MME (no explicit license declared at the time of writing), POPE (MIT License), LLaVA-Bench (Apache 2.0 License), the LLaVA model (Apache 2.0 License), and Shikra (CC-BY-NC 4.0 License). All resources are employed strictly for research purposes, and we do not redistribute them beyond what is permitted by their original licenses and terms of use.

\section{Maintaining Cognition Ability}\label{sec::cognition}

\begin{table}[t]
\centering
\scriptsize
\begin{tabular}{lccccc}
\toprule
\multicolumn{1}{c}{MME-Cognition} & \multicolumn{1}{c}{\textbf{CS}} & \multicolumn{1}{c}{\textbf{NR}} & \multicolumn{1}{c}{\textbf{Trans}} & \multicolumn{1}{c}{\textbf{CR}} & \textbf{Total*} \\
\midrule
Greedy       & 112.86 & 70.00 & 107.50 & 67.50 & \textbf{357.86}  \\
OPERA & 112.86 & 47.50 & 90.00 & 60.00 & 310.36 \\
VCD  & {120.00} & 37.50 & 97.50 & 72.50 & 327.50 \\
ICD  & {120.71} & {70.00} & 92.50 & 65.00 & 348.21 \\
SID  & {122.86} & 55.00 & 85.00 & 65.00 & 327.86 \\
VDD & 110.14 & 58.00 & 68.00 & {86.00} & 322.64 \\
M3ID & 110.86 & 61.00 & 92.50 & {75.00} & 339.36 \\
PAI & 116.43 & {70.00} & 100.00 & 67.50 & {353.93} \\
IBD & 117.14 & {70.00} & {107.50} & 57.50 & 352.14 \\
API-C & 115.00 & 47.50 & {107.50} & 47.50 & 317.50 \\
API-L & 120.00 & 45.00 & {115.00} & 45.00 & 325.00 \\
ViCrop-R & 115.00 &	50.00 &	105.00 &	65.00 &	335.00  \\
ViCrop-G & 109.29	&50.00&	105.00&	86.00&	350.29\\
\rowcolor{gray!15}
\abb{} (Ours) & {117.86} & 42.50 & {107.50} & {87.50} & \underline{355.36} \\
\bottomrule
\end{tabular} 
\caption{Comparison of different hallucination control decoding methods on the MME-Cognition subset, including Commonsense Reasoning (CS), Numerical Reasoning (NR), Translation (TR), and Code Reasoning (CR).}
\label{tab:cog_result}
\end{table}

Many hallucination control decoding methods enhance perception by increasing the contribution of visual information during token generation. While these approaches generally improve perceptual ability, they often disrupt the calibrated balance between the vision and language modalities, potentially compromising the inherent cognitive capabilities learned by the VLM during large-scale pretraining.

To investigate this, we evaluate our method and baselines on the Cognition split of MME, where answering requires reasoning beyond direct visual cues. As shown in \Cref{tab:cog_result}, most CD-based methods that explicitly modify output logits exhibit a notable decline in cognition ability, aligning with expectations. Surprisingly, PAI and IBD, which enhance the attention weight of visual tokens through intermediate embedding modifications, do not incur a substantial loss in cognition performance. This suggests that the bias introduced by their layer-wise manipulation is relatively mild and can be compensated for by subsequent model layers.
Furthermore, despite API-based methods modify the image without altering the VLM’s internal reasoning, they still lead to degraded performance. We hypothesize that the gray masking applied to images in the cognition set where backgrounds are typically uniform may introduce additional perceptual burden, ultimately impacting overall performance.

In contrast, our proposed \abb{} method does not negatively affect cognition. Although our approach is designed solely for hallucination control rather than cognition enhancement, the total cognition score remains comparable to greedy decoding and outperforms other methods, demonstrating its robustness in preserving reasoning capabilities.

\begin{figure}[t]
  \centering
    \scriptsize

    \begin{tikzpicture}
    \begin{axis}[
        xlabel={$\hm{\beta}=$},
        xlabel style={
            anchor=west,
            at={(axis description cs:0,0)},
            yshift=-1.5ex, 
            xshift=-2ex 
        },
        ylabel={MME Perception Score},
        extra y ticks={625},
        title={Perception Scores by $\alpha$ (enhance coefficient) and $\beta$ (IR threshold)},
        title style={yshift=-2ex},
        xtick={0, 0.1, 0.2, 0.3, 0.4, 0.5, 0.6, 0.7, 1.0},
        xticklabels={0, 0.1, 0.2, 0.3, 0.4, 0.5, 0.6, 0.7, $1.0{+}$},
        legend style={
        at={(0.99,0.53)}, anchor=north east,
        draw=none, fill=none,
        legend columns=2,
        /tikz/every even column/.append style={column sep=1em}
    },
        grid=major,
        width=0.95\linewidth,
        height=0.45\linewidth
    ]

    \addplot+[mark=*, thick] coordinates { (0, 680)
        (0.2, 680) (0.3, 673.33) (0.4, 665.0) (0.5, 663.33) (0.6, 665) (0.7, 665) (1.0, 665)
    };
    \addlegendentry{$\alpha$=7}

    \addplot+[mark=square*, thick] coordinates { (0, 675)
        (0.2, 675.0) (0.3, 683.33) (0.4, 670.33) (0.5, 673.33) (0.6, 675) (0.7, 680) (1.0, 665.0)
    };
    \addlegendentry{$\alpha$=10}

    \addplot+[mark=triangle*, thick] coordinates { (0, 650)
        (0.2, 653.67) (0.3, 668.33) (0.4, 663.33) (0.5, 665.0) (0.6, 680) (0.7, 685) (1.0, 680.0) 
    };
    \addlegendentry{$\alpha$=15}

    \addplot[domain=0:1.0, dotted, very thick, gray] {630};
    \addlegendentry{greedy}

    \draw[->, thick, brown, dotted]
    (axis cs:1.0,680.0) -- (axis cs:1.05,680.0);

    \draw[->, thick, red, dotted]
    (axis cs:1.0,665.0) -- (axis cs:1.05,665.0);

    \end{axis}
    \end{tikzpicture}
    
  \caption{Hyperparameter analysis of $\alpha$ and $\beta$ on MME. \abb{} demonstrates robust performance across different settings.}
  \label{fig::params}
  
\end{figure}

\section{Hyperparameter Analysis on $\alpha$ and $\beta$}

We analyze the effect of $\alpha$ and $\beta$, which control magnification strength and the stopping criterion for iterative refinement, respectively. Increasing $\alpha$ enhances heatmap contrast and strengthens magnification, while decreasing $\beta$ increases refinement steps, improving coverage but at higher computational cost (capped at 8 iterations).
 As shown in \Cref{fig::params}, smaller $\alpha$ pairs well with low $\beta$, which already ensures broad region coverage, reducing the need for strong magnification. 
Overall, the perception magnifier remains robust across a wide range of hyperparameters, consistently identifying key visual regions. Notably, using a larger $\beta$ approximates the performance of the basic \abb{} w/o iterative refinement, yet substantially mitigates hallucinations with only $2\times$ the compute.

\begin{table*}[th!]
\centering
\small
\begin{tabular}{lccccc|cccc}
\toprule
\multicolumn{1}{c}{\multirow{2}{*}{}} & \multicolumn{2}{c}{\textbf{Object-level}} & \multicolumn{2}{c}{\textbf{Attribute-level}} & \textbf{MME Score} & \multicolumn{4}{c}{\textbf{POPE Accuracy}} \\
 & \textbf{Existence}    & \textbf{Count}    & \textbf{Position}      & \textbf{Color}  & \textbf{Total*} & \textbf{AOKVQA} & \textbf{GQA} & \textbf{COCO} & \textbf{AVG*} \\
\midrule
Greedy & 190.00 & 160.00 & 71.67 & 155.00 & 576.67 & 81.87 & 81.27 & 83.46 & {82.19} \\

VCD & 185.00 & 163.33 & 65.00 & 148.33 & 561.67 & 81.86 & 81.14 & 83.31 & 82.10 \\
ICD & 195.00 & 165.00 & 65.00 & 145.00 & 575.00 & 81.93 & 81.05 & 83.78 & 82.25 \\
SID  & 195.00 & 155.00 & 73.33 & 145.00 & 568.33 & 81.57 & 81.40 & 83.43 & 82.13 \\
VDD  & 195.00 & 155.00 & 73.33 & 155.00 & 578.33 & 81.99 & 81.74 & 83.03 & 82.26 \\
M3ID  & 195.00 & 160.00 & 75.00 & 155.00 & 585.00 & 81.92 & 81.54 & 83.43 & 82.30 \\
API-C & 185.00 & 165.00 & 76.67 & 165.00 & 591.67 & 82.21 & 81.28 & 83.47 & 82.32 \\
API-L  & 190.00 & 165.00 & 73.33 & 160.00 & 588.33 & 81.84 & 81.24 & 83.33 & 82.14 \\
\rowcolor{gray!15}
\abb{} (Ours) & 190.00 & 165.00 & 68.33 & 190.00 & \textbf{613.33} & 82.15 & 81.27 & 83.66 & \textbf{82.36} \\

\bottomrule
\end{tabular} 
\caption{Evaluation results for MME Perception Scores (left) and POPE Accuracy (right) on the \textbf{Shikra} VLM. Columns marked with * indicate the overall scores for the benchmark. Some methods are omitted due to limitations on computational resources and implementations. The proposed \abb{} method consistently demonstrates superior performance on both benchmarks.}\label{tab:shikra}
\end{table*}

\section{Quantitative Results on POPE}
Table~\ref{tab:full_pm} reports detailed POPE results across 9 splits. Overall, \abb{} achieves the best performance on most splits and attains the strongest overall average, indicating that perception-guided magnification is effective for mitigating object hallucination in yes/no existence-based VQA. Notably, the absolute gain in AVG is modest for all training-free methods, because POPE accuracies with LLaVA-1.5 are already relatively high and thus largely bounded by the fixed backbone.

We observe that the \texttt{GQA-popular} splits yield smaller improvements for \abb{}. This subset tends to contain denser scenes with lower object saliency, making attention-guided magnification less decisive. In addition, \texttt{popular} questions are more susceptible to language-prior bias toward frequent object categories, where contrastive debiasing methods (e.g., VDD/M3ID) can be slightly stronger by explicitly suppressing such priors. In contrast, \abb{} primarily enhances perceptual evidence via magnification while keeping the model’s internal vision-language calibration intact. Despite this , \abb{} remains competitive on the \texttt{popular} splits and consistently improves on the remaining splits, demonstrating robust generalization across datasets and question distributions.

\section{Performance on Shikra}
To assess the generalizability of \abb{} beyond LLaVA, we apply it, along with several comparative methods, to Shikra-7B \cite{chen2023shikra}, a VLM designed for referential dialogue. The performance is evaluated on MME and POPE using Shikra with iterative refinement threshold $\beta=1.5$. The results are presented in \Cref{tab:shikra}. Note that certain comparative methods are excluded either due to the lack of available implementations for Shikra or constraints on the computational resources available.

As shown in \Cref{tab:shikra}, the proposed method continues to improve performance on Shikra compared to the baselines, though the improvement is less pronounced than on LLaVA. This discrepancy may stem from Shikra's lower visual resolution and fewer visual tokens, which impact the effectiveness of perception magnification. With fewer visual tokens, cases where an object is partially recognized at intermediate layers but not reflected in the final prediction become less frequent, as each token carries greater attention weight in the inference process. Therefore, the proposed method has fewer opportunities to enhance recognition. Additionally, the reduced number of visual tokens lowers the granularity of the token-level perception map, leading to less precise perception magnification, which may further limit the performance gains of \abb{}.

This observation suggests that the proposed method could be particularly beneficial for high-resolution VLMs, where small local features are more likely to be detected at intermediate layers but fail to significantly influence the final prediction. However, verifying this hypothesis is constrained by computational resource limitations.

Overall, the results on Shikra provide strong evidence that the proposed \abb{} method generalizes well beyond LLaVA and effectively enhances hallucination mitigation in applicable VLMs.

\section{Qualitative Samples of the PM}

Figure \ref{fig:add_qualitative}
presents additional qualitative examples illustrating the input question, the original image, the processed perception map $\mathcal{P}$, and the perception-magnified image. The results demonstrate that the proposed method effectively enlarges small objects relevant to the question while preserving fine details. In many cases, the VLM benefits from the perception-magnified image, as it can now recognize previously overlooked objects that were too small to be detected in the original image. This leads to more accurate predictions, as observed in the first and second rows.

Furthermore, when the VLM erroneously identifies a non-existent object as present, as shown in the third example, \abb{} magnifies the corresponding region, enabling the model to reassess the scene with finer detail. This process helps reduce false positive hallucinations by magnifying the visual details hence improve perceptibility. However, localized magnification does not always guarantee improved visual faithfulness. As demonstrated in the fourth row, despite providing a closer view, the model now fails to recognize the cabinet in the background. This may be due to the cabinet being identified based on its structural context rather than fine-grained textures, highlighting the limitations of region-focused magnification.

Overall, these results suggest that the proposed method effectively enhances the perceptibility of visual detail on regions crucial for next-token generation while generally preserving the overall structural integrity of objects, ultimately reducing hallucinations and improving model performance.

\section{Limitations and Future Works}\label{sec::limitations}

While the proposed \abb{} method improves VLM inference by mitigating hallucinations, magnifying image regions may also introduce certain drawbacks. This section discusses key limitations of our approach.

\begin{figure*}[t!]
    \centering
    \begin{subfigure}[b]{0.43\textwidth}
        \centering
        \includegraphics[width=\textwidth]{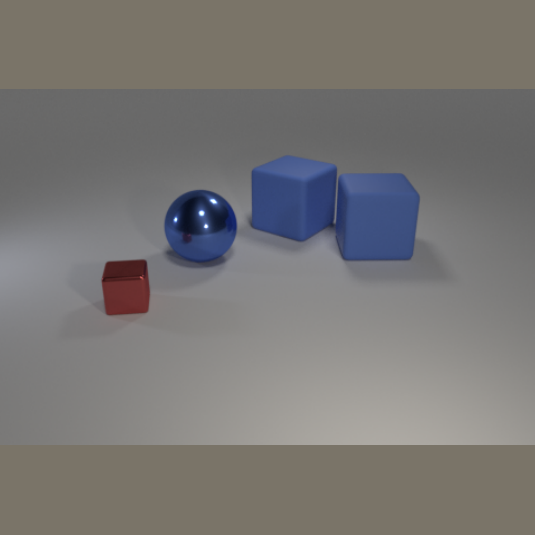}
    \end{subfigure}
    \hfill
    \begin{subfigure}[b]{0.43\textwidth}
        \centering
        \includegraphics[width=\textwidth]{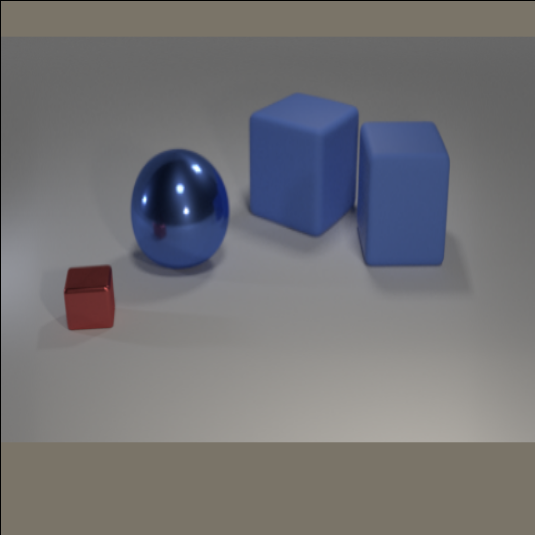}
    \end{subfigure}
    \caption{A sample CLEVR-style image (left) and its corresponding magnified visual input (right). For tasks involving shape or size-related attributes, the proposed \abb{} may introduce unintended distortions due to unregulated magnification. In this example, magnifying both the red cube and the blue sphere results in an enlarged sphere and a deformed cube, which may negatively affecting model predictions.}

    \label{fig:clvr}
\end{figure*}

\paragraph{Shape Distortion in Region Magnification.} Although region magnification allows the model to observe fine-grained visual details without altering the overall image structure, it may distort local shape properties. Specifically, in tasks requiring precise shape or length recognition, such as distinguishing letters or assessing object sizes, \abb{} does not guarantee proportional magnification. This can introduce challenges for accurate recognition and induces hallucinations. For instance, when applied to CLEVR-style questions, as shown in \Cref{fig:clvr}, with the query “Please estimate how many times large is the blue sphere compared to the red cube?”, \abb{} magnifies both objects, causing the blue sphere to appear visually larger while distorting the shape of the other two cubes. Such unintended modifications may negatively impact final predictions. Future work could explore refinements in perception map construction or post-processing techniques to achieve more structure-preserving magnification.
To address the issue of distortion, several enhancements can be integrated into our framework. One potential direction involves incorporating detected bounding boxes to constrain the perception map, thereby regularizing the magnification process and preserving object shape integrity. Another approach is to group visual tokens with shared semantic representations, enabling uniform resampling across coherent regions and reducing uneven deformation within individual objects. Furthermore, for tasks where geometric fidelity is critical such as those involving precise reasoning over shape or size, magnification can be applied adaptively at selective decoding steps rather than continuously, allowing better control over the trade-off between visual detail enhancement and structural preservation.

\begin{figure}[t!]
    \centering
    \small
    \includegraphics[width=\linewidth]{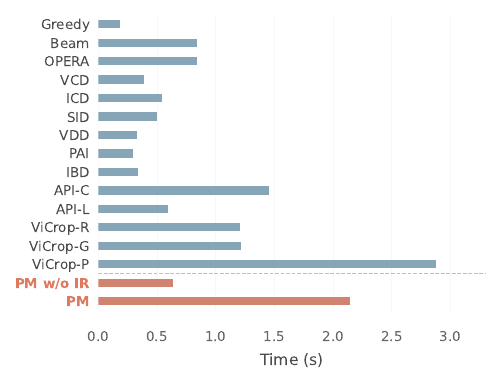}
    \caption{Average per-question running time (s) for different decoding methods.}
    \label{fig:runtime_bar}
\end{figure}

\definecolor{forestgreen}{RGB}{34, 139, 34}
\begin{figure*}[ht!]
    \centering

    \begin{minipage}[t]{0.2\textwidth}
    \vspace{-3.3cm}
    
    \begin{tabular}{@{}l@{\hspace{2pt}}r@{}} 
    \textbf{Question:} \\
    Is there a building \\ in the image? \vspace{0.2cm} \\
    
        Greedy: & \textcolor{red}{\textbf{No}} \\ 
        \abb{}: & \textcolor{forestgreen}{\textbf{Yes}}
    \end{tabular}
\end{minipage}
    \begin{minipage}[t]{0.25\textwidth}
        \centering
        \includegraphics[width=\textwidth]{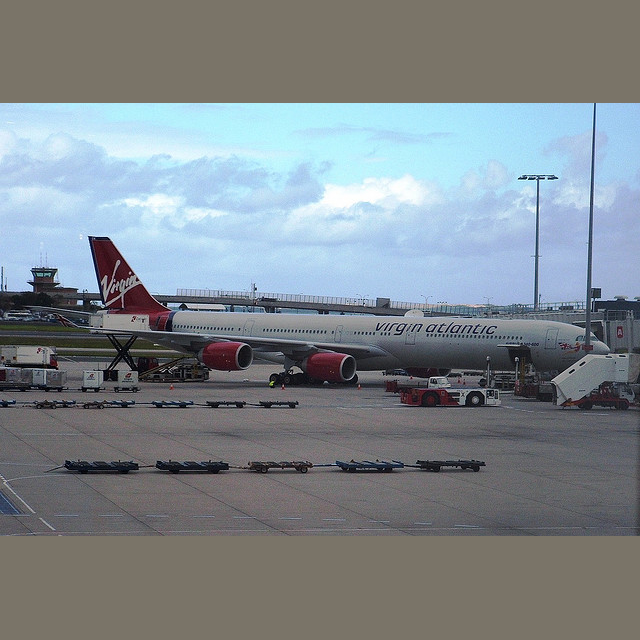} 
    \end{minipage}
    \begin{minipage}[t]{0.25\textwidth}
        \centering
        \includegraphics[width=\textwidth]{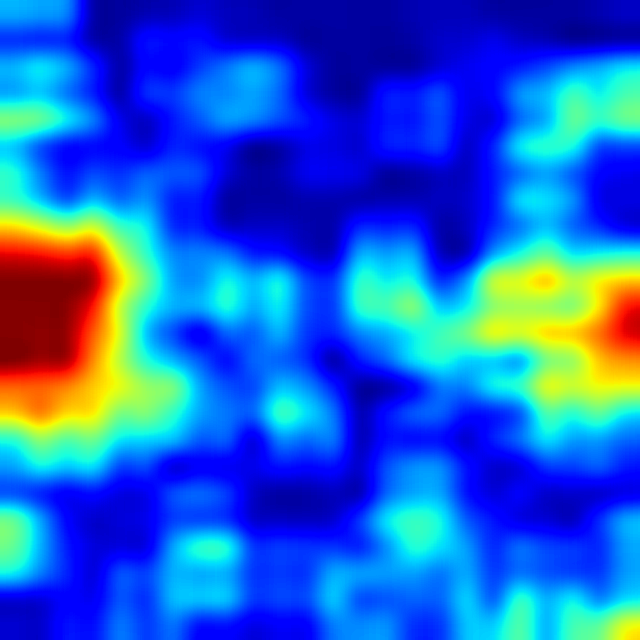}
    \end{minipage}
    \begin{minipage}[t]{0.25\textwidth}
        \centering
        \includegraphics[width=\textwidth]{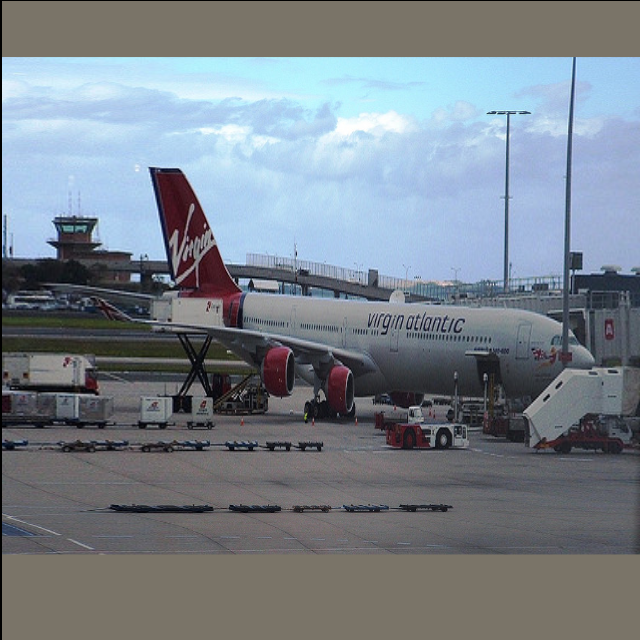}
    \end{minipage}

    \vspace{0.1cm}
    
    \begin{minipage}[t]{0.2\textwidth}
        \vspace{-3.3cm}
        
    \begin{tabular}{@{}l@{\hspace{2pt}}r@{}} 
    \textbf{Question:} \\
    Is there a bottle \\ in the image?  \vspace{0.2cm}\\
       
        Greedy: & \textcolor{red}{\textbf{No}} \\ 
        \abb{}: & \textcolor{forestgreen}{\textbf{Yes}}
    \end{tabular}
    \end{minipage}
    \begin{minipage}[t]{0.25\textwidth}
        \centering
        \includegraphics[width=\textwidth]{samples/padded_ori_1312.png} 
    \end{minipage}
    \begin{minipage}[t]{0.25\textwidth}
        \centering
        \includegraphics[width=\textwidth]{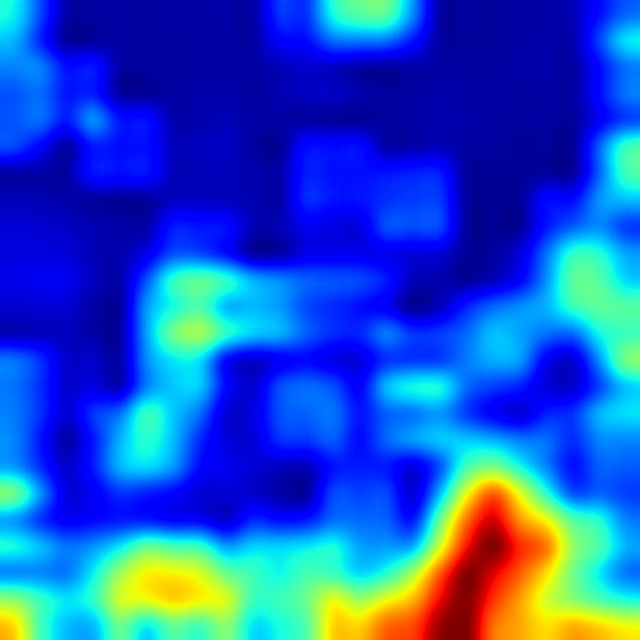}
    \end{minipage}
    \begin{minipage}[t]{0.25\textwidth}
        \centering
        \includegraphics[width=\textwidth]{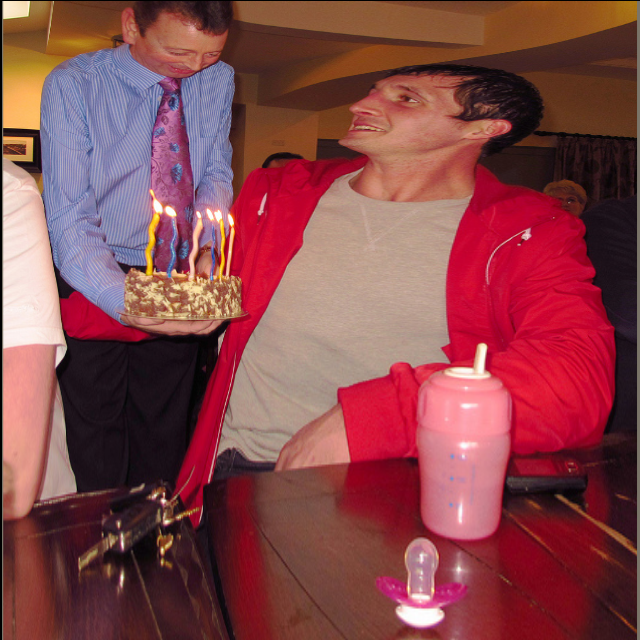}
    \end{minipage}

    \vspace{0.1cm}

    \begin{minipage}[t]{0.2\textwidth}
        \vspace{-3.3cm}
        
    \begin{tabular}{@{}l@{\hspace{2pt}}r@{}} 
    \textbf{Question:} \\
    Is there a paper \\ in the image? \vspace{0.2cm}\\
        Greedy: & \textcolor{red}{\textbf{Yes}} \\ 
        \abb{}: & \textcolor{forestgreen}{\textbf{No}}
    \end{tabular}
    \end{minipage}
    \begin{minipage}[t]{0.25\textwidth}
        \centering
        \includegraphics[width=\textwidth]{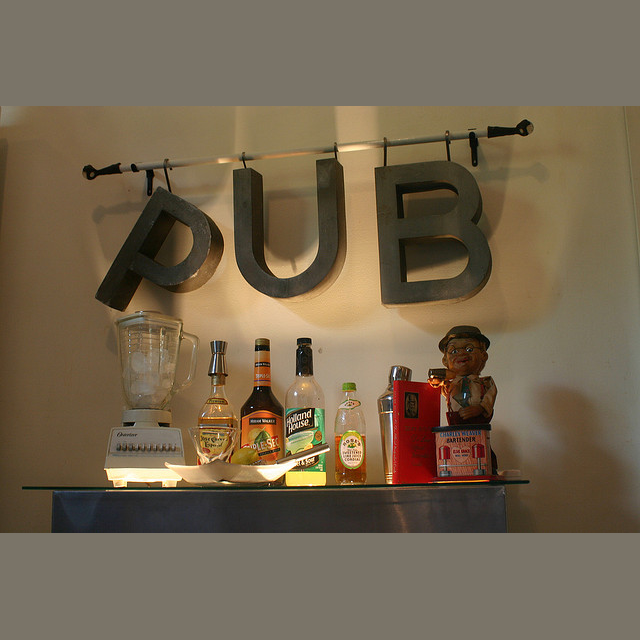} %
    \end{minipage}
    \begin{minipage}[t]{0.25\textwidth}
        \centering
        \includegraphics[width=\textwidth]{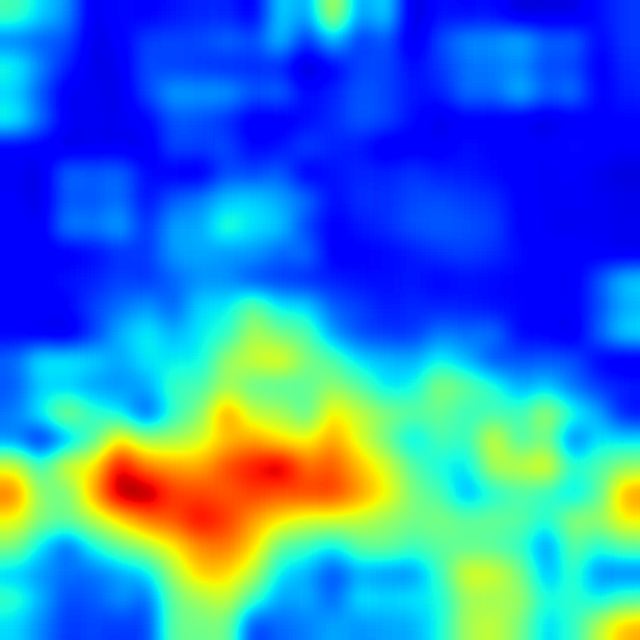}
    \end{minipage}
    \begin{minipage}[t]{0.25\textwidth}
        \centering
        \includegraphics[width=\textwidth]{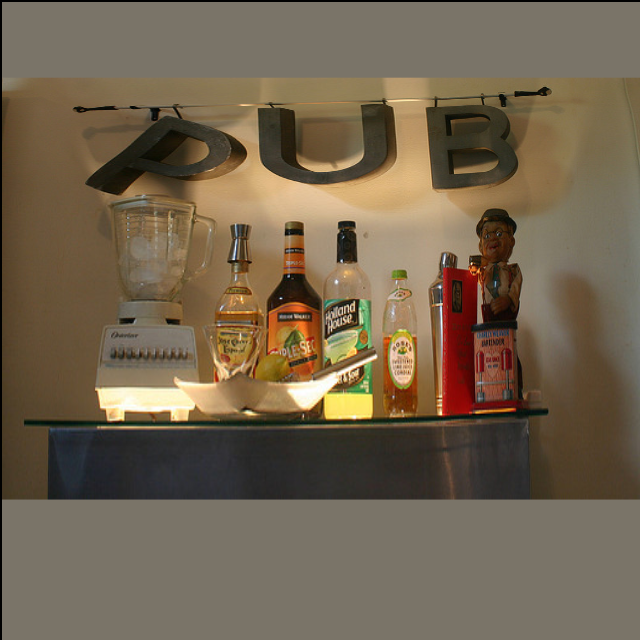}
    \end{minipage}

    \vspace{0.1cm}

    \begin{minipage}[t]{0.2\textwidth}
        \vspace{-3.3cm}
        
    \begin{tabular}{@{}l@{\hspace{2pt}}r@{}} 
    \textbf{Question:} \\
    Is there a cabinet \\ in the image? \vspace{0.2cm}\\
        
        Greedy: & \textcolor{forestgreen}{\textbf{Yes}} \\ 
        \abb{}: & \textcolor{red}{\textbf{No}}
    \end{tabular}
    \end{minipage}
    \begin{minipage}[t]{0.25\textwidth}
        \centering
        \includegraphics[width=\textwidth]{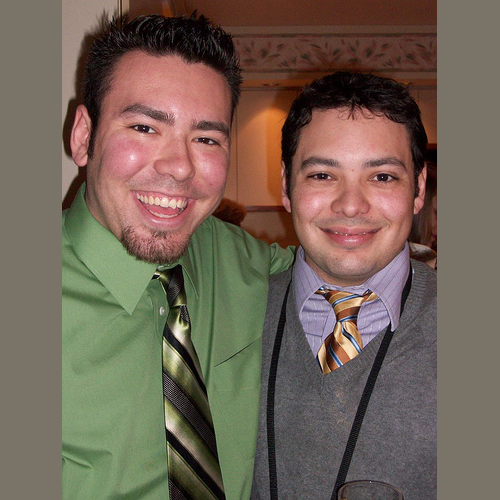} %
    \end{minipage}
    \begin{minipage}[t]{0.25\textwidth}
        \centering
        \includegraphics[width=\textwidth]{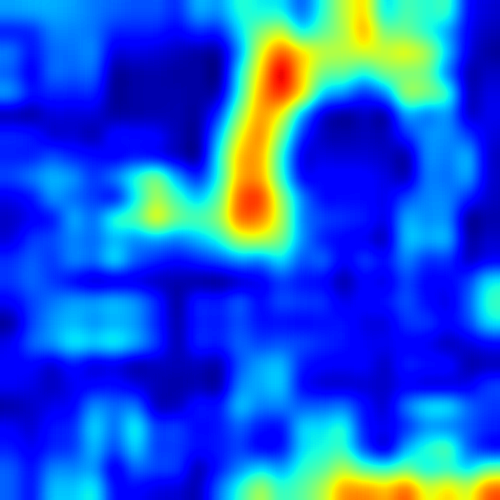}
    \end{minipage}
    \begin{minipage}[t]{0.25\textwidth}
        \centering
        \includegraphics[width=\textwidth]{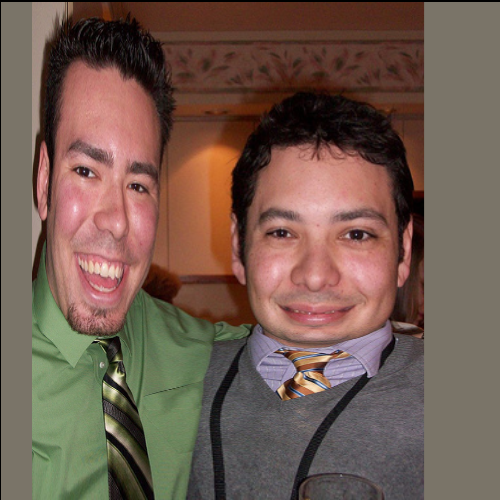}
    \end{minipage}

    \caption{Additional qualitative examples on the POPE benchmark showing the input question, the original image, the processed perception map $\mathcal{P}$, and the perception-magnified image, arranged from left to right. The answers generated by greedy decoding and \abb{} are compared, with correct answers highlighted in green and incorrect ones in red.}\label{fig:add_qualitative}
\end{figure*}

\paragraph{Efficiency.}

A key limitation of our method lies in its computational overhead: applying perception-based magnification at every decoding step requires feature extraction from a modified image, which precludes the use of efficient VLM decoding techniques such as key-value (KV) caching. 
As depicted in \Cref{fig:runtime_bar}, \abb{} without Iterative Refinement (IR) is approximately $3.5\times$ slower than greedy decoding, while \abb{} with IR increases this to $\sim 12\times$. In comparison, contrastive-decoding baselines such as IBD and VDD incur only $\sim 2$--$3\times$ inference time, whereas prompting-based methods with multi-pass processing (e.g., API and ViCrop) typically cost $8$--$20\times$ depending on the variant. These results suggest that \abb{} exposes a controllable cost--performance trade-off: \abb{} w/o IR already yields substantial gains at moderate overhead, while full \abb{} can be reserved for latency-tolerant settings.

Nonetheless, our method remains relatively efficient among perception-oriented approaches when used without iterative refinement, as the perception map can be constructed with only a single additional VLM forward pass. To further mitigate runtime cost, \abb{} can be applied adaptively only at decoding steps where visual information is critical (e.g., during object description), while being omitted for tokens with lower visual dependency such as punctuation or purely reasoning-oriented connectives. A practical indicator for such adaptation is the aggregated attention mass on visual tokens, which signals when the model is actively grounding its predictions in the image. Finally, future work may explore speculative or deferred refinement strategies, where magnification is applied post hoc to selectively revise previously decoded spans, thereby reducing the frequency of image resampling and improving scalability for longer generations.

\paragraph{Improving Compatibility.}
While the proposed method can be easily applied to interleaved VLMs with visual tokens corresponding to all patch features, applying it to VLMs where the correlation between visual tokens and image regions are non-trivial, can be slightly more complicated. The method depends on certain mapping from the visual token attention weights to the attention span on the original image. 
Therefore, when apply \abb{} to models where the visual tokens are constructed with not only a feature extractor, but also a token sampler or cross-attention feature extractors, \abb{} will need some attention mapping mechanism between visual tokens and patches. However, we can resolve this problem by introducing gradient backpropagation to address the implicit relationships between the image and tokens, refining the visual attentions and construct the perception map, similar to method proposed in \cite{map1}. Overall, such limitation does not reduce the contribution of this work, as the idea of adaptive-magnification can also always be used in aligned with interpretability tools designed for different VLMs. Alternatively, one could use \abb{} in ways similar to \cite{api}, where any compatible model could be used as external guidance to the decoding-adaptive magnification.

\begin{table*}[t!]\centering

\begin{minipage}{0.99\linewidth}\vspace{0mm}    \centering
\begin{tcolorbox}
    \centering
    \small
     \hspace{-6mm}
    \begin{tabular}{p{0.99\linewidth}}

\tbf{Prompt for Reference-based Evaluation}\\
\texttt{[}Context\texttt{]}\\
An aerial view of Diamond Head in the Hawaiian Islands.\\
\\
\texttt{[}Question\texttt{]}\\
What is the name of this famous sight in the photo?\\
\\
\texttt{[}Assistant 1\texttt{]}\\
The famous sight in the photo is Diamond Head.\\
\\
\texttt{[}End of Assistant 1\texttt{]}\\
\\
\texttt{[}Assistant 2\texttt{]}\\
The famous sight in the photo is the Na Pali Coast, which is a scenic coastal area on the island of Kaua'i in Hawaii.\\
\\
\texttt{[}End of Assistant 2\texttt{]}\\
\\
\texttt{[}System\texttt{]}\\
We would like to request your feedback on the performance of two AI assistants in response to the user question displayed above. The user asks the question on observing an image. For your reference, the visual content in the image is represented with a few sentences describing the image. \\ Please rate the helpfulness, relevance, accuracy, level of details of their responses. Each assistant receives an overall score on a scale of 1 to 10, where a higher score indicates better overall performance.\\ Please first output a single line containing only two values indicating the scores for Assistant 1 and 2, respectively. The two scores are separated by a space.\\ In the subsequent line, please provide a comprehensive explanation of your evaluation, avoiding any potential bias and ensuring that the order in which the responses were presented does not affect your judgement.

\\\\

\midrule

\tbf{Response:} 
\\
10 2\\
\\
Assistant 1 correctly identified the famous sight in the photo as Diamond Head, which is consistent with the context provided. The response is accurate, relevant, and directly answers the question with the appropriate level of detail given the context.\\\\
Assistant 2, on the other hand, incorrectly identified the sight as the Na Pali Coast, which is not mentioned in the context. This response is inaccurate and irrelevant to the question, as it does not align with the description of the image provided. Additionally, the details about the Na Pali Coast, while informative, are not applicable to the context of Diamond Head. Therefore, Assistant 2's response is not helpful in this scenario.

    \end{tabular}
\end{tcolorbox}
    
\vspace{-2mm}
\end{minipage}\caption{Prompt for reference-based evaluation.}
    \label{tab:ref_eval}
\end{table*}

\section{GPT-Assisted Evaluation}\label{sec::app_gpt}
We adopt the GPT-assisted evaluation prompts directly from prior studies. For {reference-based evaluation}, we follow the prompt used in \cite{llava}, with an example of the prompt and corresponding evaluation result provided in \Cref{tab:ref_eval}. For {vision-based evaluation}, we use the prompt from \cite{opera}, with an example shown in \Cref{tab:v_acc_eval}.

\subsection*{Evaluation Protocol}

Each tested method is evaluated under two settings:

\begin{itemize}
    \item \textbf{Reference-based Evaluation}: GPT-4o is given the question along with a detailed textual description of the image.
    \item \textbf{Vision-based Evaluation}: GPT-4o is given the question and the original image.
\end{itemize}

In both settings, GPT-4o scores two answers: one from the tested method and another from GPT-4 (used as the reference). Each answer is rated independently on a scale of 0 to 10. The final evaluation score is computed as:
\[
\text{Relative Score} = \frac{\text{Score}_{\text{method}}}{\text{Score}_{\text{GPT-4}}}
\]
This normalization ensures comparability across questions of varying difficulty.

\subsection*{Evaluation Criteria}

The evaluation criteria differ slightly between the two settings:

\begin{itemize}
    \item In \textbf{reference-based evaluation}, GPT-4o assesses overall content quality, coherence, and informativeness of the response relative to the image description.
    \item In \textbf{vision-based evaluation}, GPT-4o is explicitly instructed to focus on the \textit{accuracy} of visual grounding and the \textit{level of detail} present in the answer.
\end{itemize}

\section{LLM Usage Statement}
We used GPT-4.5 exclusively during the writing phase of this paper, for the purpose of polishing the manuscript. Specifically, we provided prompts such as: ``Revise the grammar and academic wording of this paragraph, and list aspects to be improved, including suggestions on how to improve them.'' All outputs were subsequently reviewed and manually revised to ensure that technical accuracy was preserved. No part of the research ideation or experimental analysis relied on LLM assistance.

\begin{table*}[ht!]\centering
\begin{minipage}{0.99\linewidth}\vspace{0mm}    \centering
\begin{tcolorbox}
    \centering
    \small
     \hspace{-6mm}
    \begin{tabular}{p{0.99\linewidth}}

\tbf{Prompt for Vision-based Evaluation}
\\
\begin{minipage}[c]{.99\textwidth}
  \centering
\includegraphics[width=0.145\linewidth]{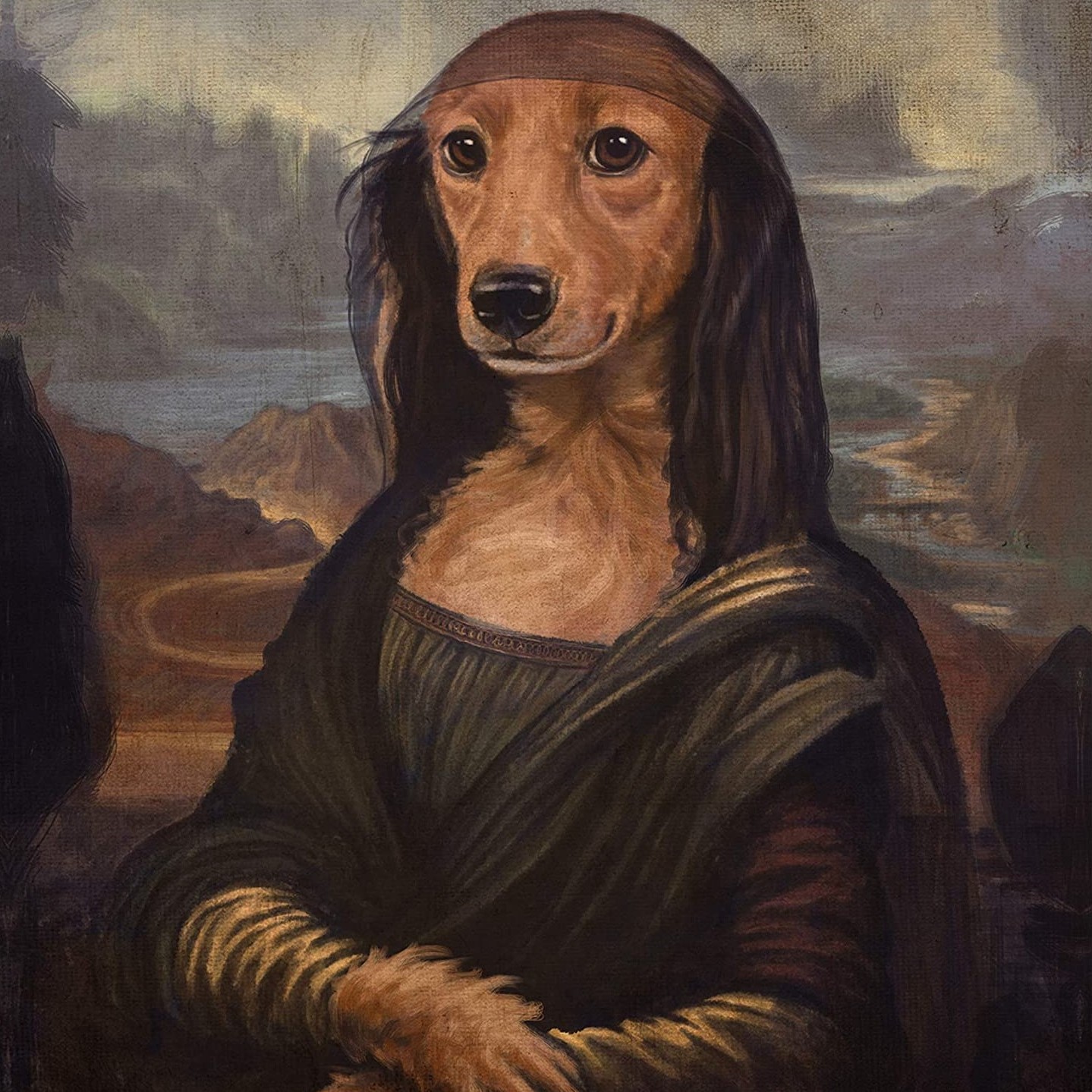}
\end{minipage} \\
You are required to score the performance of two AI assistants in describing a given image. You should pay extra attention to the hallucination, which refers to the part of descriptions that are inconsistent with the image content, such as claiming the existence of something not present in the image or describing incorrectly in terms of the counts, positions, or colors of objects in the image. Please rate the responses of the assistants on a scale of 1 to 10, where a higher score indicates better performance, according to the following criteria:\\
1: Accuracy: whether the response is accurate with respect to the image content. Responses with fewer hallucinations should be given higher scores.\\
2: Detailedness: whether the response is rich in necessary details. Note that hallucinated descriptions should not count as necessary details.\\
Please output the scores for each criterion, containing only two values indicating the scores for Assistant 1 and 2, respectively. The two scores are separated by a space. Following the scores, please provide an explanation of your evaluation, avoiding any potential bias and ensuring that the order in which the responses were presented does not affect your judgement.\\
\\
\texttt{[}Assistant 1\texttt{]}\\
The intended effect of this painting might be to create a humorous and whimsical twist on the famous Mona Lisa artwork by featuring a dog dressed in her attire, which can evoke feelings of amusement and curiosity.\\
\texttt{[}End of Assistant 1\texttt{]}\\
\\
\texttt{[}Assistant 2\texttt{]}\\
The intended effect of this painting is to create a humorous and whimsical scene by combining a dog with a human-like appearance, such as a Renaissance-style painting. The dog is wearing a Renaissance-style dress and a hat, which adds a touch of humor and creativity to the image. The painting might be intended to evoke amusement and curiosity in the viewer, as it is not common to see a dog dressed in such a manner. The artist might have wanted to challenge the viewer's expectations and create a unique, memorable piece of art.\\
\texttt{[}End of Assistant 2\texttt{]}\\
\\
Output format:\\
Accuracy: \textless Scores of the two answers\textgreater\\
Reason:\\
\\
Detailedness: \textless Scores of the two answers\textgreater\\
Reason: \\

\midrule

\tbf{Response:} 
\\
``9": **Accuracy: 8 9**\\\\**Reason:** \\
Assistant 1 accurately describes the image as a humorous twist on the Mona Lisa featuring a dog, but lacks specific details about the attire. Assistant 2 provides a more detailed and accurate description, mentioning the Renaissance-style dress and the humorous intent, which aligns well with the image content.\\\\
**Detailedness: 6 9**
\\\\**Reason:** 
\\
Assistant 1 gives a brief description focusing on the humorous aspect but lacks detail about the attire and style. Assistant 2 offers a richer description, noting the Renaissance-style dress and hat, and elaborates on the artistic intent, providing a more comprehensive understanding of the image."

    \end{tabular}
\end{tcolorbox}
    
\vspace{-2mm}
\end{minipage}\caption{Prompt for vision-based evaluation, note the image is processed by GPT-4o for evaluation.}
    \label{tab:v_acc_eval}
\end{table*}

\end{document}